%% file: paper.tex
\RequirePackage{fix-cm}
\documentclass[12pt]{article}

\def\smartqed{\def\qed{\ifmmode\squareforqed\else{\unskip\nobreak\hfil
\penalty50\hskip1em\null\nobreak\hfil\squareforqed
\parfillskip=0pt\finalhyphendemerits=0\endgraf}\fi}}
\smartqed  

\usepackage{bm}
\usepackage{bbm}
\usepackage{xspace}
\usepackage{color}
\usepackage{amsmath, amsfonts, amssymb, amsthm}
\usepackage{enumitem}
\usepackage{hyperref}
\usepackage{algorithm}
\usepackage{algpseudocode}
\usepackage{graphicx}
\usepackage{natbib}

\usepackage[labelformat=simple]{subcaption}
\renewcommand\thesubfigure{(\alph{subfigure})}

\usepackage[margin=1in]{geometry}

\input{headers}

\newtheorem{theorem}{Theorem}
\newtheorem{corollary}{Corollary}

\begin{document}

\title{Robust Linear Classification from Limited Training Data\footnote{Accepted for publication in the Machine Learning Journal.}}

\author{Deepayan Chakrabarti\\
           McCombs School of Business\\
           University of Texas\\
           Austin, TX, USA\\
           {\tt deepay@utexas.edu}}
\date{}

\maketitle

\begin{abstract}
\input{abstract}
\end{abstract}

\input{intro.tex}
\input{main}
\input{alg}

\input{exp}
\input{related}

\input{conc}

\section*{Code availability}
The code is available from \url{https://faculty.mccombs.utexas.edu/deepayan.chakrabarti/mywww/software/ROLIN-2021.tgz}.






\bibliographystyle{spbasic}
\bibliography{paper}

%
%
%
%
%
%
%

\appendix

\input{appendix}

\end{document}

%% file: headers.tex
\newcommand{\Ourmethod}{\textsc{RoLin}\xspace}
\newcommand{\RobustCV}{\textsc{RobustCV}\xspace}
\newcommand{\CVoneSD}{CV-1-SD\xspace}

\newcommand{\bx}{{\ensuremath{\bm x}}\xspace}
\newcommand{\bz}{{\ensuremath{\bm z}}\xspace}
\newcommand{\bb}{{\ensuremath{\bm \beta}}\xspace}
\newcommand{\bbw}{\ensuremath{{\bm \beta}_{w}}\xspace}

\newcommand{\bbSz}{\ensuremath{{\bm \beta}_{\mathcal{S}_0}}\xspace}
\newcommand{\bbSot}{\ensuremath{{\bm \beta}_{\mathcal{S}_1\cup\mathcal{S}_2}}\xspace}

\newcommand{\gbx}{{\ensuremath{g_\bb(\bx)}}\xspace}

\newcommand{\Szero}{\ensuremath{{\mathcal{S}_0}}\xspace}
\newcommand{\Sone}{\ensuremath{{\mathcal{S}_1}}\xspace}
\newcommand{\Stwo}{\ensuremath{{\mathcal{S}_2}}\xspace}
\newcommand{\Sot}{\ensuremath{{\mathcal{S}_1\cup\mathcal{S}_2}}\xspace}
\newcommand{\Si}{\ensuremath{{\mathcal{S}_i}}\xspace}

\newcommand{\Vzero}{\ensuremath{V_{\Szero}}\xspace}
\newcommand{\Vone}{\ensuremath{V_{\Sone}}\xspace}
\newcommand{\Vtwo}{\ensuremath{V_{\Stwo}}\xspace}
\newcommand{\Vot}{\ensuremath{V_{\Sot}}\xspace}
\newcommand{\Vi}{\ensuremath{V_{\Si}}\xspace}

\newcommand{\bone}{{\ensuremath{\bm 1}}\xspace}

\DeclareMathOperator*{\argmin}{arg\,min}

\newcommand{\txtred}[1]{\textcolor{red}{#1}}
\renewcommand{\txtred}[1]{#1}

%% file: abstract.tex
We consider the problem of linear classification under general loss functions in the limited-data setting.
Overfitting is a common problem here.
The standard approaches to prevent overfitting are dimensionality reduction and regularization.
But dimensionality reduction loses information, while regularization requires the user to choose a norm, or a prior, or a distance metric.
We propose an algorithm called \Ourmethod that needs no user choice and applies to a large class of loss functions.
\Ourmethod combines ``reliable'' information from the top principal components with a robust optimization to extract any useful information from ``unreliable'' subspaces.
It also includes a new robust cross-validation that is better than existing cross-validation methods in the limited-data setting. 
Experiments on $25$ real-world datasets and three standard loss functions show that \Ourmethod broadly outperforms both dimensionality reduction and regularization.
Dimensionality reduction has $14\%-40\%$ worse test loss on average as compared to \Ourmethod.
Against $L_1$ and $L_2$ regularization, \Ourmethod can be up to 3x better for logistic loss and 12x better for squared hinge loss.
The differences are greatest for small sample sizes, where \Ourmethod achieves the best loss on 2x to 3x more datasets than any competing method.
For some datasets, \Ourmethod with $15$ training samples is better than the best norm-based regularization with $1500$ samples.

%% file: intro.tex
\section{Introduction}
\label{sec:intro}


In many machine learning applications, the size of the training data is small relative to the number of features, and acquiring more data may be too costly or time-consuming.
For example, a standard dataset to predict breast-cancer from gene-expression data has only $99$ positive examples for $7,650$ features~\citep{sotiriou_breast_2003}, and datasets with smaller training sizes are also of interest~\citep{blagus_smote_2013}.
In computational advertising, we must learn to predict whether an ad is relevant to a person after seeing only limited data for that ad.
Since there are many available ads, waiting for more training data can reduce ad revenue.
Limited training data also leads to the ``cold-start'' problem in recommendation systems, where we must quickly tune our recommendations for new users, or find relevant matches for new items.
Thus, the limited-data setting is widely applicable.

We consider the problem of learning a linear classifier from limited training data.
\txtred{In such cases, overfitting is common.
In other words, the feature weight vector that minimizes the training loss often has a test loss that is much worse than the training loss.}
The usual solution for such overfitting is to do dimensionality reduction or regularization.
Dimensionality reduction reduces the number of features, while regularization keeps all features but penalizes large weights.
For example, for the least-squares loss, Principal Components Regression (PCR) does dimensionality reduction, while Ridge and LASSO do regularization based on the $L_2$ and $L_1$ norms respectively.

However, both dimensionality reduction and regularization have weaknesses.
Dimensionality reduction ignores information.
For example, PCR uses only the top few principal components of the feature matrix.
But, if the top principal components are uncorrelated with the response variable, the PCR solution may perform poorly~\citep{jolliffe1982note}.
For regularization, it is not easy to choose the best $L_p$ norm.
It depends on the dataset, the training size, and the loss function.
Different norms can lead to significantly different test losses.
Existing explanations of regularization rely on priors, or a distance between probability distributions, or a distance metric in feature space.
It is not clear why such inputs are needed and how we should choose them in practice.
This motivates the following problem:

\medskip
{\em How can we build a linear classifier that (a) outperforms both dimensionality reduction and norm-based regularization, (b) works for a wide range of loss functions, and (c) needs no user input such as a norm or a prior?
}



\begin{figure}[tbp]
  \begin{center}
	\includegraphics[width=\textwidth]{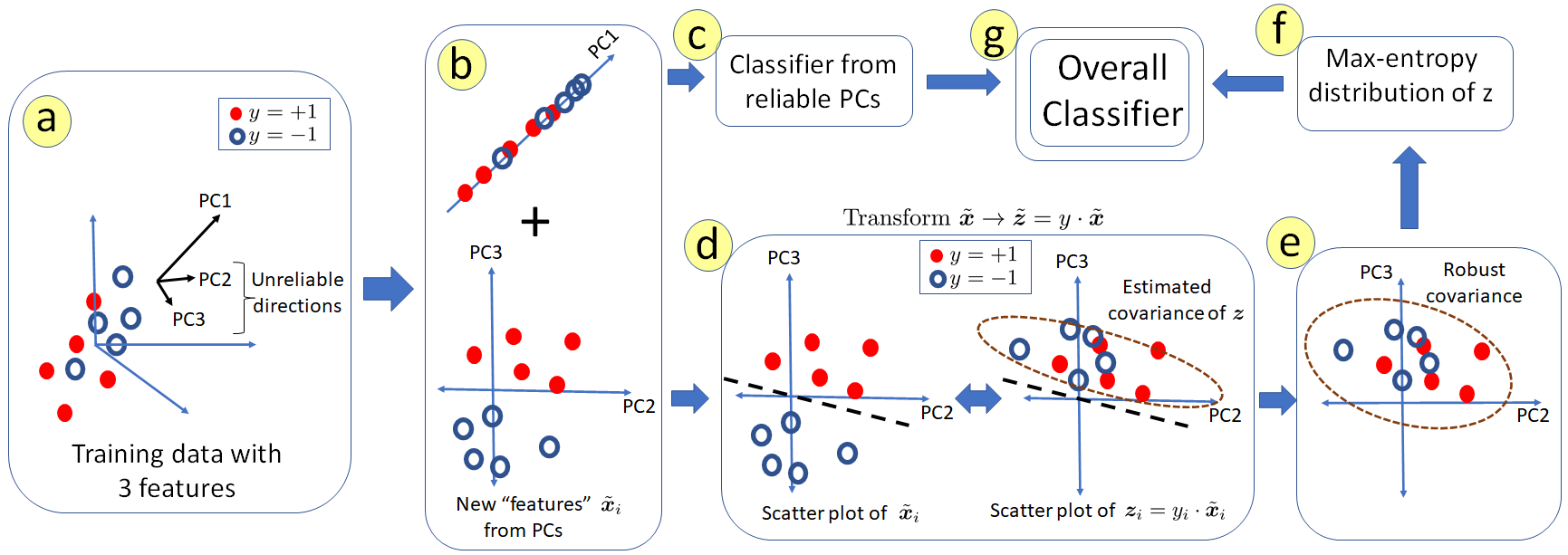}
	\caption{\footnotesize{\txtred{{\em Overview of \Ourmethod:} 
(a) We can reliably estimate the top Principal Components (PCs) of the data distribution from limited training data.
In this example, only the first PC is reliably estimated.
(b) We project the data onto the reliable and unreliable subspaces.
(c) We find the classifier with the least training loss on data projected on the reliable subspace.
(d) But for the orthogonal subspace, such a classifier may overfit.
Linear classifiers such as logistic regression minimize a loss that is a function of ${\bm z}:=y\cdot{\tilde{\bm x}}$, where $\tilde{\bm x}$ is a projected datapoint.
Given limited training data, even the covariance estimate of ${\bm z}$ can be noisy~\citep{marcenko_distribution_1967}.
So, minimizing the loss over the empirical distribution of ${\bm z}$ can yield classifiers (shown by the dashed line) that have low training loss but much higher test loss.
(e) \Ourmethod builds a robust covariance of ${\bm z}$.
(f) It uses this to construct a maximum-uncertainty distribution for ${\bm z}$.
(g) Optimizing on this distribution gives a robust classifier, which \Ourmethod combines with the reliable classifier from step (c).
	}}}
  \label{fig:hilevel}
	\end{center}
\end{figure}

\medskip
Our proposed method, called \Ourmethod (RObust LINear classification), aims to achieve this by combining dimensionality reduction with robust optimization.
Dimensionality reduction methods such as PCR rely on the idea that solutions constructed from the top singular vectors (principal components) are less prone to overfitting.
Like PCR, \Ourmethod first constructs such a classifier.
But, unlike PCR, we do not ignore the bottom singular vectors.
Training and test loss can indeed be very different for data projected on to the bottom singular vectors.
But even from this ``unreliable'' projected data, we may be able to estimate some low-order moments, such as the mean and some aspects of the covariance.
Now, the loss function depends on the entire data distribution, not just the low-order moments.
So, \Ourmethod constructs a ``worst-case'' distribution that matches the low-order moments.
Then it finds a classifier that has the smallest loss under this distribution.
Now, we have two classifiers: one from the top principal components, and a robust one from the rest.
\Ourmethod combines them into a single classifier that captures all available information.
Thus, \Ourmethod goes beyond the top few principal components, but still avoids overfitting.
Figure~\ref{fig:hilevel} shows the intuition behind \Ourmethod.

\smallskip
We summarize our main contributions below.


\smallskip\noindent{\em A new approach to avoid overfitting:}
Given limited training data, the top principal components are often reliable while the bottom components are noisy.
Motivated by this, \Ourmethod processes the top and bottom principal components differently.
Reliable data from the top components is used directly, while noisy data from the bottom components is filtered through a robust optimization.
This ensures that all available reliable information is extracted and used in building \Ourmethod's classifier.
By limiting the robust optimization to the unreliable subspace, \Ourmethod avoids becoming too conservative.





\smallskip\noindent{\em No user choice needed:} 
In contrast to existing regularization methods, \Ourmethod does not force the user to choose a norm, prior, or distance metric.

\smallskip\noindent{\em Applicable to many loss functions:}
\Ourmethod works, unchanged, for the logistic, hinge, squared hinge, and modified Huber losses, among others.
In particular, we can use \Ourmethod for both logistic regression and linear SVMs.

\smallskip\noindent{\em Robust cross-validation:}
Existing cross-validation methods identify overfitting classifiers by their poor accuracy on holdout sets.
But in limited-data settings, holdout sets are small and holdout accuracy may be too noisy.
We develop a new cross-validation method, called \RobustCV, that checks for several signs of overfitting that are missed by standard cross-validation.



\smallskip\noindent{\em Empirical results:}
We compare \Ourmethod against competing methods on three loss functions and $25$ real-world datasets, where the number of features ranges from $p=8$ to $p=43,680$.
We test each dataset under five different training sizes, from $n=15$ to $n=200$ samples.
\Ourmethod outperforms dimensionality reduction as well as $L_1$ and $L_2$ regularization.
Dimensionality reduction has $14\%-40\%$ worse loss on average that \Ourmethod, under all problem settings.
For some datasets, dimensionality reduction can be 4x worse.
Under logistic loss, \Ourmethod can be up to 3x better than the best norm-based regularization.
Under squared hinge loss, \Ourmethod can be up to 12x better.
\Ourmethod performs particularly well for small training sizes, where robustness is most important.
When $50$ or fewer training samples are available, \Ourmethod achieves the smallest loss on around 2x to 3x as many datasets as the next best method, depending on the loss function.
For some datasets, \Ourmethod with $n=15$ samples is better than both $L_1$ and $L_2$ regularization with $n=1500$ samples.
Finally, among the competitors of \Ourmethod, no single method dominates, and it is challenging to choose the best method for a given dataset, loss function, and training size.
In contrast, we find that \Ourmethod works well for all datasets under all problem settings.



The rest of the paper is organized as follows.
We present our robust formulation and the main theorems in Section~\ref{sec:robust}.
Section~\ref{sec:algCV} provides detailed algorithms for \Ourmethod and \RobustCV.
Section~\ref{sec:exp} presents empirical results.
We discuss prior work on overfitting in Section~\ref{sec:related}, and we conclude in Section~\ref{sec:conc}.
All proofs are deferred to Appendix~\ref{app:theorem}.

%% file: main.tex
\section{Robust Minimization of Expected Loss}
\label{sec:robust}

We are given $n$ independent training samples from some distribution $\cal{D}$ of pairs $(\bx, y)\in\mathbb{R}^p\times \{-1, 1\}$, where $\bx$ is a feature vector with $p$ features, and $y$ is a binary class label.
We want to train a classifier, parameterized by \bb, to output a positive score $\gbx$ when it predicts $y=1$, and a negative score otherwise.
The quality of classification is measured by a loss function $\ell(y, \gbx)$.
The best classifier is the one that minimizes the expected loss
\begin{align}
\min_{\bb} E_{(y,\bx)\in\cal{D}}\; \ell(y, \gbx).\label{eq:minExpLoss}
\end{align}
We consider linear classifiers where $\gbx=\beta_0 + \bbw^T \bx$, where $\beta_0$ (intercept) and $\bbw$ (feature weights) are the first and the remaining elements of $\bb\in\mathbb{R}^{p+1}$.
In this setting, many common losses are functions of $y\cdot\gbx$, and we denote the loss $\ell(y, \gbx)$ as $\ell(y\cdot\gbx)$ henceforth.
Common loss functions include
\begin{align}
\ell(y\cdot\gbx) & = \left\{
	\begin{array}{cl}
	\log_2(1+\exp(-y\cdot\gbx)) & \text{(logistic loss)}\\
	\max(0, 1-y\cdot\gbx) & \text{(hinge loss)}\\
	\left(\max(0, 1-y\cdot\gbx)\right)^2 & \text{(squared hinge loss)}\\
	\mathbbm{1}_{y\gbx\geq -1}\cdot\max(0, 1-y\cdot\gbx)^2 - \mathbbm{1}_{y\gbx < -1}\cdot (4y\cdot\gbx) & \text{(modified Huber loss)}\\
	\mathbbm{1}_{y\cdot\gbx \leq 0} & \text{(zero-one loss).}
	\end{array}\right.
  \label{eq:lossfunctions}
\end{align}
Well-known classifiers such as logistic regression (logistic loss) and linear SVM (hinge or squared hinge loss) fall under this framework.
Such linear classifiers are also the building blocks for popular complex classifiers such as neural networks.
Except for zero-one loss, all the other losses are convex in \bb.
In this paper, we seek to minimize the expected loss in Eq.~\ref{eq:minExpLoss} for such convex loss functions.



\Ourmethod splits this problem into separate problems in different subspaces of the feature space.
Next, we discuss the details of subspace separation, our robust optimization, and its solution.
But first, we discuss the connection to PCR in more detail, as this helps us explain the unique features of \Ourmethod.

\subsection{Intuition for Subspaces via PCR}
\label{sec:robust:intuition}
Consider the problem of least-squares regression:
\begin{align}
\min_{\bb\in\mathbb{R}^p} E_{\bx, y} (y-\bb^T \bx)^2 =
\min_{\bb\in\mathbb{R}^p} E\left[y^2\right] - 2\bb^T E\left[y\cdot\bx\right] + \bb^T E\left[ \bx\bx^T\right]\bb,
\label{eq:LS}
\end{align}
where we assume a zero intercept for ease of exposition.
Suppose we are given $n$ i.i.d. training samples $(y_i, \bx_i)\in\mathbb{R}\times\mathbb{R}^p$.
Then we can solve Eq.~\ref{eq:LS} after replacing the expectation terms with their estimates.
But for small $n$, estimation errors lead to poor out-of-sample performance.
Instead, Principal Components Regression (PCR) first projects the features $\bx_i$ on to the top few principal components.
Then, it solves Eq.~\ref{eq:LS} only on the projected data.
In other words, PCR splits the feature space $\mathbb{R}^p$ into a subspace $\Sone$ spanned by the top principal components, and the orthogonal subspace $\Stwo$.
It then solves for the best $\bb\in\Sone$ and ignores $\Stwo$.

The reason for the success of PCR is as follows.
The principal directions and singular values correspond to the eigenvectors and eigenvalues of the matrix $\hat{M} = \sum_i \bx_i\bx_i^T/n$.
The top eigenvalues and eigenvectors of $\hat{M}$ are often close to those of the expectation matrix $M=E[\hat{M}]=E[\bx\bx^T]$, {\em even for small training sizes}.
This is because the estimation error for an eigenvector depends on the gap between its eigenvalue and all other eigenvalues~\citep{davis1970rotation, yu_useful_2015}.
A larger gap implies smaller estimation error.
For many datasets, this gap is large for the top eigenvalues.
So the top principal directions are well estimated, and the same holds for the singular values too~\citep{zhao19portfolio}.
Hence, $\hat{M}$ and $M$ have similar projections on the subspace \Sone spanned by these well-estimated principal directions.
So for any $\bb\in\Sone$, $\bb^T \hat{M}$ is close to $\bb^T M$, and so $\bb^T \hat{M} \bb \approx \bb^T M \bb$.
Applying this in Eq.~\ref{eq:LS}, $\min_{\bb\in\Sone} E(y-\bb^Tx)^2 \approx \min_{\bb\in\Sone} \sum_i (y_i - \bb^T \bx_i)^2 / n$, and this becomes PCR's solution.
In contrast, the remaining principal components are poorly estimated when $n$ is small.
So, the least-squares training loss is not a reliable indicator of the expected loss in the orthogonal subspace \Stwo.
Therefore PCR ignores \Stwo.

The loss functions we consider in Eq.~\ref{eq:lossfunctions} are not restricted to just second moments as in Eq.~\ref{eq:LS}.
However, the basic ideas underlying PCR are still applicable, as we discuss next.

\subsection{Subspace Separation}
\label{sec:robust:subspace}
We want to minimize the expected loss under a linear score function.
The loss is given by $\ell(y\cdot\gbx) = \ell(y\beta_0 + \bbw^T(y\cdot\bx)) = \ell(y\beta_0 + \bbw^T \bz)$, where $\bb=(\beta_0\; \bbw)^T$ with intercept $\beta_0\in\mathbb{R}$ and feature vector $\bbw\in\mathbb{R}^p$, and $\bz=y\cdot\bx$.
Extending the PCR argument, we propose to split the space $\mathbb{R}^p$ into three subspaces \Szero, \Sone, and \Stwo. 
The subspace $\Szero$ is spanned by the top few eigenvectors of $\hat{M} = \sum_i \bx_i\bx_i^T/n$.
We expect that the loss function can be reliably estimated in this subspace: 
\begin{align}
E\left[\ell(y\beta_0 + \bbw^T\bz)\right] 
  \approx \mathbb{P}_n \left[\ell(y\beta_0 + \bbw^T\bz)\right] 
	\quad\text{for any $\bbw\in\Szero$,}
	\label{eq:expect0}
\end{align}
where $\mathbb{P}_n[.]$ represents the empirical mean.
The next few eigenvectors span the subspace $\Sone$.
Here,  we can reliably estimate only the first and second moments of the distribution of $\bz$ projected on to $\Sone$:
\begin{align}
\bbw^T E\left[\bz\bz^T\right] \bbw 
  \approx \bbw^T \mathbb{P}_n\left[\bz \bz^T\right] \bbw 
	\quad\text{for any $\bbw\in\Sone$}
	\label{eq:expect1}
\end{align}
But this may not be enough to estimate the loss function accurately.
We must also consider the subspace $\Stwo$ that is orthogonal to both $\Szero$ and $\Sone$.
Here, we only expect first moments to be well-estimated.
For the subspace $\Stwo$, orthogonal to both $\Szero$ and $\Sone$, only first moments are well-estimated.
\begin{align}
\bbw^T E\left[\bz\right] 
  \approx \bbw^T \mathbb{P}_n \left[\bz\right] 
	\quad\text{for any $\bbw\in\Sot$}
	\label{eq:expect2}
\end{align}
But the second moments under \Stwo are not arbitrary.
Note that \Sone and \Stwo are constructed from separate sets of eigenvectors of the sample covariance matrix $\mathbb{P}_n[\bz\bz^T] = \mathbb{P}_n[\bx\bx^T] = \hat{M}$ \txtred{(since $\bz=y\cdot\bx$ and $y\in\{+1, -1\}$).}
So they are orthogonal under $\hat{M}$, that is,
$\mathbb{P}_n\left[ \left(P_{\Sone} \bz\right)\left(P_{\Stwo} \bz\right)^T \right] = 0$,
where $P_{\Si}\in\mathbb{R}^{p\times p}$ is a matrix that projects any vector on to $\Si$, for $i\in\{0, 1, 2\}$.
We expect \Sone and \Stwo to remain nearly orthogonal under the population covariance $M$:
\begin{align}
E\left[ \left(P_{\Sone} \bz\right)\left(P_{\Stwo} \bz\right)^T \right] 
  &\approx 0.
	\label{eq:expect3}
\end{align}
Finally, we expect the eigenvalues of the second-moment matrix under \Stwo to be smaller than those under \Sone.
\begin{align}
\sigma_{max}\left( E\left[ \left(P_{\Stwo} \bz\right)\left(P_{\Stwo} \bz\right)^T \right] \right)
  &< \sigma_{min} \left( \mathbb{P}_n\left[ \left(P_{\Sone} \bz\right)\left(P_{\Sone} \bz\right)^T \right] \right),
	\label{eq:expect4}
\end{align}
where $\sigma_{max}$ and $\sigma_{min}$ refer to the maximum and minimum non-zero eigenvalues of a matrix.

\subsection{Different Optimizations over Subspaces}
\label{sec:robust:opt}
To construct our solution, we first optimize the training loss over $\Szero$.
By Eq.~\ref{eq:expect0}, the training loss accurately reflects the expected loss of a solution \bb with intercept $\beta_0$ and $\bbw\in\Szero$.
So, we choose an intercept $\beta_0\in\mathbb{R}$ and $\bbSz\in\Szero$ that solves the following optimization:
\begin{align}
\min_{\beta_0\in\mathbb{R}, \bbSz\in\Szero} \frac{1}{n} \sum_{i=1}^n \ell\left(\beta_0\cdot y_i + \bbSz^t (P_{\Szero} \bz_i)\right)
  = \min_{\beta_0, \bbSz} \frac{1}{n} \sum_{i=1}^n \ell\left(y_i \cdot \left(\beta_0 + \bbSz^t (P_{\Szero} \bx_i)\right)\right)
\label{eq:bbSz}
\end{align}
This is the usual parameter-fitting problem in classification but with projected features  $P_\Szero\bx_i$.
We can use off-the-shelf solvers for logistic regression (for logistic loss) or linear SVM (hinge or squared hinge losses). 
For other convex loss functions, such as the modified Huber loss, we can use standard optimizers such as stochastic gradient descent.

Now, unlike PCR, we do not ignore $\Sot$.
Suppose we set $\bbw=\bbSz + \bbSot$ for some vector $\bbSot\in\Sot$.
Then, the expected loss $E[\ell(\beta_0\cdot y + \bbw^T\bz)]$ equals $E[ \ell\left(\beta_0\cdot y + \bbSz^T (P_\Szero \bz) + \bbSot^T (P_\Sot \bz)\right)]$.
If we change \bbSot, it affects the third term but not the first two.
\txtred{Setting $\bbSot=0$ corresponds to dimensionality reduction, because we only use the top principal components in \Szero.
But a careful choice of \bbSot can reduce the loss obtained from \bbSz alone.}
But we cannot just project the data on to \Sot and pick the \bbSot that minimizes training loss.
This is because we cannot reliably estimate the loss function in this subspace.
Instead, we need a \bbSot that is robust to estimation errors.
To avoid being too conservative, we still need to  use all available information about \Sot (Eqs.~\ref{eq:expect1}--\ref{eq:expect4}). 
We formulate this as a robust optimization problem, which we discuss next.

\subsection{Robust Formulation}
\label{sec:robust:robust}
To select a robust \bbSot, we need to characterize the distribution of the data projected on to \Sot. 
The empirical distribution is unreliable here.
Instead, we will construct distributions that are ``worst-case'', in that they have the maximum uncertainty subject to the constraints in Eqs.~\ref{eq:expect1}--\ref{eq:expect4}.
Then, we pick the \bbSot with the best worst-case performance.
This prevents \bbSot from overfitting to incidental aspects of the empirical distribution, while still using all reliable information about moments.

We note that our worse-case distribution depends on the data, but not on the weight vector \bbSot.
An alternative notion of robustness is to let the worst-case distribution depend on \bbSot as well.
This corresponds to setting, for each possible choice of \bbSot, the worst possible higher-order moments of the data distribution.
Since our only constraints are on the mean and covariance, setting all other moments to their worst-case values is overly conservative.
Fixing the data distribution to the maximum-uncertainty distribution helps us achieve robustness in a more practical way.

We will formulate our robust model assuming that Eqs.~\ref{eq:expect1}--\ref{eq:expect3} are equalities.
The first moments of the distribution of $P_\Sot\bz$ can be taken to be the first moments of the empirical distribution.
For the second moments of this distribution, we have only partial information.
Let \Vot be a matrix whose columns are the eigenvectors that span \Sone and \Stwo.
Writing the second-moment matrix of $P_\Sot\bz$ in the basis \Vot, we get a block-wise form:
\begin{multline}
\Vot^T E\left[\left(P_{\Sot}\bz\right)\left(P_{\Sot}\bz\right)^T\right] \Vot
  \\=\begin{bmatrix} 
      \begin{array}{c@{\hspace{1em}}c}
        \Vone^T E\left[\left(P_{\Sone}\bz\right)\left(P_{\Sone}\bz\right)^T\right] \Vone &
        \Vone^T E\left[\left(P_{\Sone}\bz\right)\left(P_{\Stwo}\bz\right)^T\right] \Vtwo\\[2ex]
        \Vtwo^T E\left[\left(P_{\Stwo}\bz\right)\left(P_{\Sone}\bz\right)^T\right] \Vone &
        \Vtwo^T E\left[\left(P_{\Stwo}\bz\right)\left(P_{\Stwo}\bz\right)^T\right] \Vtwo
      \end{array}
    \end{bmatrix} 
  =: \begin{bmatrix} B_{11} & B_{12}\\ B_{21} & B_{22} \end{bmatrix} 
\label{eq:second}
\end{multline}
Now,
\begin{align*}
B_{11} = \Vone^T E \left[ \left(P_{\Sone}\bz\right) \left(P_{\Sone}\bz\right)^T \right] \Vone = \Vone^T \mathbb{P}_n \left[ \left(P_{\Sone}\bz\right) \left(P_{\Sone}\bz\right)^T \right] \Vone = \mathbb{P}_n \left[ \left(\Vone^T\bz\right) \left(\Vone^T\bz\right)^T\right],
\end{align*}
where the second equality follows from applications of Eq.~\ref{eq:expect1}, and the third equality follows from $P_\Sone=\Vone\Vone^T$. 
So $B_{11}$ is the second-moment matrix of the data projected on to \Vone.
Also, by Eq.~\ref{eq:expect3},
\begin{align}
B_{12} = B_{21} = 0.
\label{eq:B12}
\end{align}
For $B_{22}$, we have no estimates but only a bound (Eq.~\ref{eq:expect4}).
This suggests the following uncertainty set for $B_{22}$:
\begin{align}
B_{22} \in \mathcal{U} := \left\{W\left| W\succeq \frac{1}{n} \sum_{i=1}^n \left(\Vtwo^T\bz_i\right)\left(\Vtwo^T\bz_i\right)^T, \|W\| \leq \sigma_{bound}\right.\right\},
\label{eq:B22}
\end{align}
where $\sigma_{bound} = \sigma_{min} \left( \mathbb{P}_n\left[ \left(P_{\Sone} \bz\right)\left(P_{\Sone} \bz\right)^T \right] \right)$.
Note that by construction, this uncertainty set is non-empty.
Equations~\ref{eq:second}--\ref{eq:B22} thus characterize the second moments of the distribution of $P_\Sot\bz$.


Now, we construct our worst-case distribution for $P_\Sot\bz$; call it $q(P_\Sot\bz)$.
We choose $q(P_\Sot\bz)$ to be the distribution with the maximum entropy (and hence the most ``uncertainty'') subject to the first and second moments specified above.
It is well known that the maximum entropy is achieved by the exponential family distribution with those moments~\citep{cover06elements}.
Now, we pick \bbSot that performs best under $q(P_\Sot\bz)$:
\begin{align}
\bbSot = \argmin_{{\bm b}\in\Sot} \max_{B_{22}\in \mathcal{U}} \frac{1}{n} \sum_{i=1}^n E_{{\bm r} \sim q(.)} \left[ \ell\left(\beta_0\cdot y_i + \bbSz^T \left(P_{\Szero} \bz_i\right) + {\bm b}^T {\bm r}\right) \right],
\label{eq:bbSot}
\end{align}
where ${\bm r}$ is a random variable that represents $P_{\Sot}\bz$, and $\beta_0$ and \bbSz are the solutions of Eq.~\ref{eq:bbSz}.
Note that $q(.)$ depends on $B_{22}$.


\subsection{The Solution of the Robust Objective}
\label{sec:robust:solution}
The solution \bbSot of Eq.~\ref{eq:bbSot} depends not only on the distribution $q(P_{\Sot}\bz)$ but also on $\beta_0$, \bbSz, and $P_{\Szero}\bz_i$.
This suggests that solving the robust optimization might be difficult.
However, we show a surprising result.
While the scale of \bbSot indeed depends on all the above factors, the direction of \bbSot does not.
In fact, in many cases, {\bf the direction does not even depend on the specific loss function.}
\begin{theorem}[Direction of the robust solution]
Suppose the loss function $\ell(.)$ is non-negative, monotonically non-increasing, convex, differentiable, and the absolute value of its first derivative $|\ell'(.)|$ has finite non-zero expectation under the standard Normal distribution.
Then, any solution of Eq.~\ref{eq:bbSot} satisfies
\begin{align}
\bbSot &= c\cdot \Vot \Sigma^{-1} \mu,\label{eq:bbSotDir}\\
\text{where } \Sigma &=
    \begin{bmatrix} 
      \begin{array}{c@{\hspace{1em}}c}
        \frac{1}{n} \sum_{i=1}^n \left(\Vone^T\bz_i\right)\left(\Vone^T\bz_i\right)^T &
        0 \\[2ex]
        0 &
        \sigma_{bound}\cdot I
      \end{array}
    \end{bmatrix},\nonumber\\
{\bm \mu} &= \frac{1}{n}\sum_{i=1}^n \Vot^T\bz_i,\nonumber
\end{align}
for some scalar $c$.
Further, if there is a sequence of loss functions $\ell^{(m)}(.)$ satisfying the properties mentioned above such that $\lim_{m\to\infty}\sup_{x\in\mathbb{R}} |\ell^{(m)}(x) - \ell(x)| = 0$, then there is a solution of the form of Eq.~\ref{eq:bbSotDir} that is arbitrarily close to the optimal.
\label{thm:main}
\end{theorem}

\begin{corollary}[Wide applicability]
The minimizer \bbSot of Eq.~\ref{eq:bbSot} has the form of Eq.~\ref{eq:bbSotDir} for logistic, hinge, squared hinge, and modified Huber losses.
\label{cor:losses}
\end{corollary}

These results are significant from both a theoretical and practical standpoint.
It is challenging to formulate tractable robust optimizations.
Uncertainty sets are often chosen for their ease of analysis.
So, it is encouraging to see a simple closed-form structure emerge from a well-motivated formulation.
Further, we do not need separate analyses for each loss function.
Armed with Theorem~\ref{thm:main}, we only need to pick a single scalar, which is  the magnitude $\|\bbSot\|$.
We will choose this by cross-validation.

Computing the direction of \bbSot is also easy because $\Sigma$ is a diagonal matrix.
To see this, let $X$ be the matrix with $\bx_i$ as its $i^{th}$ row, and let $X=UDV^T$ be its singular value decomposition (SVD).
The SVD of $X$ is related to the eigenvectors and eigenvalues of $\hat{M}$ by the formula $n\cdot\hat{M} = \sum_i \bz_i\bz_i^T = \sum_i \bx_i\bx_i^T = V D^2 V^T$.
For $i\in\{0, 1, 2\}$, let $D_\Si$ be the diagonal matrix of singular values corresponding to the eigenvectors in \Vi.
Then, the top-left block of $\Sigma$ equals $\sum_i (\Vone^T\bz_i)(\Vone^T\bz_i)^T/n = D_{\Sone}^2/n$.
So, $\Sigma$ is a diagonal matrix with entries $D_{\Sone}^2/n$ and $\sigma_{bound}$.
Since $\sigma_{bound}=\min(D_\Sone^2/n)>\max(D_\Stwo^2/n)$ from Eq.~\ref{eq:expect4}, we may write $\Sigma = \text{diag}(\max(D_\Sot^2/n, \sigma_{bound}))$.
We propose using a smooth upper-bound of this: $\Sigma_{smooth} = \text{diag}(D_\Sot^2/n + \sigma_{bound})$.
By varying $\sigma_{bound}$, we get smooth transitions between different choices for \Sone and \Stwo.
Using $\Sigma_{smooth}$ also reveals a curious connection between our robust solution and ridge regression.

\begin{theorem}[Connection to ridge regression]
The robust solution \bbSot using $\Sigma_{smooth}$ is also the solution, up to a scaling factor, for regressing $y_i$ on $P_\Sot\bx_i$ with a ridge penalty:
\begin{align*}
\bbSot &\propto \argmin_{\bm b} \sum_{i=1}^n \left(y_i-{\bm b}^TP_\Sot\bx_i\right)^2 + n\sigma_{bound}\cdot \|{\bm b}\|^2.
\end{align*}
\label{thm:ridge}
\end{theorem}
Thus, we can view \Ourmethod as a mix of standard classification over the well-estimated top principal components and ridge regression over the poorly-estimated orthogonal subspace.

%% file: alg.tex
\section{Algorithm and Robust Cross-validation}
\label{sec:algCV}

\begin{algorithm}[tbp]
  \caption{Calculate \bb for \Ourmethod.}
  \label{alg:calcBeta}
  \begin{algorithmic}[1]
  \small{
  \Function{CalcBeta}{$\{\bx_i\in\mathbb{R}^p, y_i\in\{1, -1\}\mid i=1, \ldots, n\}, k, \sigma_{ratio}, b_{max}$}
    \State $\bz_i\gets y_i\cdot \bx_i$
    \State $Z \gets n\times p$ matrix whose $i^{th}$ row is $\bz_i$
    \State $U, D, V \gets \text{SVD}(Z)$
      \Comment{$Z = UDV^T, D_{i,i} \text{ in descending order}$}\label{alg:beta:SVD}
    \State $\Vzero \gets$ first $k$ columns of $V$  \Comment{Basis vectors for subspace \Szero}\label{alg:beta:Vzero}
    \State $\Vot \gets$ last $p-k$ columns of $V$\label{alg:beta:Vot}
    \State $D_{\Sot} \gets$ diagonal matrix with entries $D_{k+1,k+1}, D_{k+2,k+2}, \ldots, D_{p,p}$
    \State $\beta_0, {\bm \gamma} \gets \argmin_{\beta_0\in\mathbb{R}, {\bm \gamma}\in\mathbb{R}^k} \sum_{i=1}^n \ell\left(\beta_0\cdot y_i + {\bm \gamma}^T \Vzero^T\bz_i\right)$\label{alg:beta:bbSz1}
    \State $\bbSz \gets \Vzero{\bm \gamma}$\label{alg:beta:bbSz2}
    \State $\sigma_{bound} \gets \sigma_{ratio}\cdot \max(D_{\Sot}^2)$
    \State ${\bm \nu} \gets \Vot \left(D_{\Sot}^2 + \sigma_{bound}\cdot I\right)^{-1} \Vot^T Z^T \bone$\label{alg:beta:dir1}
    \If {$\|{\bm \nu}\| \neq 0$}
    	\State ${\bm \eta}_\Sot \gets \Vot{\bm \nu} / \|{\bm \nu}\|$\label{alg:beta:dir2}
      \State $\|\bbSot\| \gets \argmin_{c\in[0, b_{max}]} \sum_{i=1}^n \ell\left(\beta_0\cdot y_i + \left(\bbSz + c\cdot{\bm \eta}_\Sot\right)^T \bz_i\right)$\label{alg:beta:mag}
      \State $\bbSot \gets \|\bbSot\|\cdot {\bm \eta}_\Sot$
    \Else
      \State $\bbSot\gets 0$
    \EndIf
    \State $\bb \gets \text{intercept } \beta_0 \text{ and feature weights } \bbSz+\bbSot$
    \State \Return \bb
  \EndFunction
  }
  \end{algorithmic}
\end{algorithm}

\Ourmethod combines two algorithms: the \textproc{CalcBeta} algorithm to calculate the solution \bb, and the \textproc{RobustCV} algorithm to robustly select model parameters for \textproc{CalcBeta}.
We now provide details for both these algorithms.

\paragraph{Calculation of the solution vector \bb.}
Algorithm~\ref{alg:calcBeta} shows the steps in calculating \bb.
Apart from the data itself, it requires three inputs.
This first input is the number $k$ of top principal components that comprise the subspace \Szero.
The second input is a parameter $\sigma_{ratio}$ from which we construct $\sigma_{bound}$.
Tuning $\sigma_{ratio}$ allows for smooth transitions between \Sone and \Stwo. 
Setting $\sigma_{ratio}=0$ corresponds to setting $\Stwo=\varnothing$, while a large $\sigma_{ratio}$ corresponds to $\Sone=\varnothing$.
Third, we need an upper bound  $b_{max}$ on the magnitude $\|\bbSot\|$.

We first construct the matrix $Z$ with rows $\bz_i = y_i\cdot \bx_i$.
The singular value decomposition of $Z$ gives the diagonal matrix $D$ of singular values and the matrix $V$ of singular vectors (step~\ref{alg:beta:SVD})\footnote{The matrix $Z$ has the same singular values/vectors as the matrix $X$ with rows $\bx_i$.}.
We form \Vzero from the first $k$ singular vectors and \Vot from the remaining singular vectors (steps~\ref{alg:beta:Vzero}-\ref{alg:beta:Vot}).
\Vzero and \Vot span the subspaces \Szero and \Sot, respectively.
For \Szero, we compute the optimal intercept $\beta_0$ and weight vector $\bbSz\in\Szero$ via Eq.~\ref{eq:bbSz} (steps~\ref{alg:beta:bbSz1}-\ref{alg:beta:bbSz2}).
As discussed in Section~\ref{sec:robust:opt}, this step can use any convex minimizer.
Then, for \Sot, we compute the direction vector ${\bm \eta}_{\Sot}$ using Theorem~\ref{thm:main} (steps~\ref{alg:beta:dir1}-\ref{alg:beta:dir2}).
Here, we use $\Sigma_{smooth}$ with $\sigma_{bound} = \sigma_{ratio} * \max(D_{\Sot}^2)$, where $D_{\Sot}$ contains singular values corresponding to \Vot.
Finally, we choose the best magnitude of \bbSot over the training samples, but under the bound $\|\bbSot\|\leq b_{max}$ (step~\ref{alg:beta:mag}).
Bounded norm solutions have small generalization error (Section~\ref{sec:related}), so this is appropriate for the poorly-estimated subspace \Sot.
Note that the question of the ``right'' norm does not arise.
We must bound the $L_2$-norm $\|\bbSot\|$ since we already know the direction of the vector \bbSot.

\begin{algorithm}[tbp]
  \caption{Robust Cross-validation for \Ourmethod.}
  \label{alg:RobustCV}
  \begin{algorithmic}[1]
  \small{
  \Function{RobustCV}{dataset $\mathcal{D}$, $\Theta_{ratio}, \Theta_{slack}, \Theta_{gain}$}
    \State Split $\mathcal{D}$ into multiple train/holdout splits $\{(\mathcal{D}^{tr}_j, \mathcal{D}^{ho}_j)\mid j=1, \ldots, J\}$
    \State $C_\sigma, C_b \gets $ fixed set of choices for $\sigma_{ratio}$ and $b_{max}$
			\Comment{Initialization}

		\State $k_{max} \gets \text{\textproc{MaxReliablePCs}}(\Theta_{ratio})$
      \Comment{Find the maximum number of reliable PCs} 
			\label{alg:RCV:kmax}
    
    \State $\Psi_\Szero \gets \{\psi=(k, 0, 0) \mid k\in\{1, \ldots, k_{max}\}\}$
      \label{alg:RCV:PsiSzero}
    \State $\psi_\Szero^{rob} \gets \text{\textproc{RobustParams}}(\Psi_\Szero, \Theta_{slack})$
      \Comment{Robust solution using only top PCs}\label{alg:RCV:rob1}

    \State $\Psi \gets \{\psi=(k, \sigma_{ratio}, b_{max}) \mid k\in\{1, \ldots, k_{max}\}, \sigma_{ratio}\in C_\sigma, b_{max}\in C_b\}$
      \label{alg:RCV:Psi}
    \State $\psi^{rob} \gets \text{\textproc{RobustParams}}(\Psi, \Theta_{slack})$
      \Comment{Robust solution for general setting}\label{alg:RCV:rob2}


    \State $\psi^{best} \gets \psi_\Szero^{rob} \text{ if } \text{cost}(\psi^{rob}) \geq \left(1 - \Theta_{gain}\right) \text{cost}(\psi_\Szero^{rob}) \text{ else } \psi^{rob}$
			\label{alg:RCV:rob3}
    \State \Return $\psi^{best}$
  \EndFunction
  \Statex
  \Function{MaxReliablePCs}{$\Theta_{ratio}$}
    \ForAll{$k\in\{1, \ldots, p\}$}
      \State $\psi \gets (k, \sigma_{ratio}=0, b_{max}=0)$
      \State $\text{loss}_{ratio}(\psi), \text{cost}(\psi) \gets \text{\textproc{CalcCost}}(\{(\mathcal{D}^{tr}_j, \mathcal{D}^{ho}_j)\}, \psi)$
    \EndFor
    \State $k_{max} \gets \max\left(\left\{k \mid \text{loss}_{ratio}(k, 0, 0) \leq \Theta_{ratio} \quad \forall m\leq k\right\}\right)$
		\State \Return $k_{max}$
	\EndFunction
  \Statex
  \Function{RobustParams}{$\Psi$, $\Theta_{slack}$}
    \ForAll{$\psi\in\Psi$}
      \State $\text{loss}_{ratio}(\psi), \text{cost}(\psi) \gets \text{\textproc{CalcCost}}(\{(\mathcal{D}^{tr}_j, \mathcal{D}^{ho}_j)\}, \psi)$
    \EndFor
    \State $\psi^\star \gets \argmin_{\psi\in\Psi} \text{cost}(\psi)$
      \label{alg:RCVHelper2:start}
    \State $\Psi_{slack} \gets \{\psi\in\Psi \mid \text{cost}(\psi) \leq \left(1+\Theta_{slack}\right) \text{cost}(\psi^\star)\}$
      \Comment{Nearly-min-cost candidates}
    \State $\psi^{rob} \gets \argmin_{\psi\in\Psi_{slack}} \left( \text{cost}(\psi) + \text{loss}_{max}(\psi)\right)$
      \Comment{Most robust candidate}
      \label{alg:RCVHelper2:end}
    \State \Return $\psi^{rob}$
  \EndFunction
  \Statex
  \Function{CalcCost}{data splits $\left\{(\mathcal{D}_j^{tr}, \mathcal{D}_j^{ho})\mid j=1, \ldots, J\right\}$, $\psi=(k, \sigma_{ratio}, b_{max})$}
    \ForAll{$(\mathcal{D}_j^{tr}, \mathcal{D}_j^{ho})$}
      \State $\bb\gets \text{CalcBeta}\left(\mathcal{D}^{(tr)}_j, \psi\right)$
      \State $\text{loss}_j^{tr}, \text{loss}_j^{ho} \gets $ training and holdout loss using \bb
    \EndFor
    \State $\text{loss}_{avg} \gets \frac{1}{J}\sum_{j=1}^J \text{loss}_j^{ho}$
    \State $\text{loss}_{max} \gets \max_j \text{loss}_j^{ho}$
    \State $\text{loss}_{ratio} \gets \frac{1}{J}\sum_{j=1}^J\frac{\text{loss}_j^{ho}}{\text{loss}_j^{tr}}$
      \label{alg:RCVHelper:lossratio}
    \State $\text{cost} \gets \text{loss}_{avg} \text{ if } \text{loss}_{ratio} \leq \Theta_{ratio} \text{ else } \text{loss}_{max}$ 
      \label{alg:RCVHelper:robcost}
      \Comment{Robust cost}
    \State \Return $\text{loss}_{ratio}, \text{cost}$
  \EndFunction
  }
  \end{algorithmic}
\end{algorithm}

\paragraph{Robust cross-validation for choosing model parameters.}
Now we need to select the input parameters $\psi=(k, \sigma_{ratio}, b_{max})$ for \textproc{CalcBeta} (Algorithm~\ref{alg:calcBeta}).
A poor $\psi$ leads to an overconfident classifier.
But, there may only be a few holdout samples where an overconfident classifier incurs significant losses.
The averaging step of cross-validation can hide these few large losses.
To counter this, we develop a new robust cross-validation method called \RobustCV (Algorithm~\ref{alg:RobustCV}).

\RobustCV guards against overconfidence by using three signals.
The first signal is the loss ratio, which we define as the ratio of the holdout loss to training loss, averaged over all cross-validation splits (step~\ref{alg:RCVHelper:lossratio}).
A $\psi$ with a high loss ratio indicates that overfitting is likely.
To implement this idea, we use a loss ratio threshold $\Theta_{ratio}$.
Below the threshold, we use the average holdout loss as a measure of the cost of $\psi$, like standard cross-validation.
But above the threshold, the cost of $\psi$ is set to the maximum holdout loss (step~\ref{alg:RCVHelper:robcost}).
We also use the loss ratio to find an upper bound $k_{max}$ for the number of principal components that are well-estimated (step~\ref{alg:RCV:kmax}).
Throughout our algorithm, we restrict the parameter $k$ in $\psi$ to $k\leq k_{max}$.

The second warning sign of overconfidence is a significant difference between the average holdout loss and the maximum holdout loss.
Standard cross-validation picks the $\psi^\star$ with the smallest cost.
But, we find a robust parameter setting $\psi^{rob}$ whose cost is within a factor $(1+\Theta_{slack})$ of $\psi^\star$, but whose worst-case holdout loss is better (steps~\ref{alg:RCVHelper2:start}-\ref{alg:RCVHelper2:end}).

Third, realizing that the distribution of $P_\Sot\bz$ is poorly estimated, we check if a solution constructed from \Szero alone is good enough.
That is, we select a $\psi_\Szero^{rob}$ that ignores the subspace \Sot unless the cost of improves by a factor of $\Theta_{gain}$ using $\psi^{rob}$  (step~\ref{alg:RCV:rob3}).
Together, these steps ensure that \RobustCV selects a $\psi^{best}$ that is robust but not too conservative.
\Ourmethod then runs \textproc{CalcBeta} under $\psi^{best}$ over the entire training sample.
The output of this is \Ourmethod's solution.

%% file: exp.tex
\section{Experiments}
\label{sec:exp}

We will first compare \Ourmethod against competing methods for $25$ real-world datasets and three loss functions.
Then, we will contrast \Ourmethod run with \RobustCV versus alternative cross-validation schemes.
Finally, we will present a sensitivity analysis for \Ourmethod's parameters. 

\paragraph{Datasets.}
We use $25$ benchmark real-world datasets from the UCI repository\footnote{\url{https://archive.ics.uci.edu/ml/datasets}}.
These span many domains, and the number of features range for $p=8$ to $p=43,680$.
We convert categorical variables  into binary ``dummy'' variables, and count each dummy variable as a separate feature.

\paragraph{Evaluation methodology.}
We run experiments with logistic loss, squared hinge loss, and modified Huber loss (Eq.~\ref{eq:lossfunctions}).
All three are standard loss functions, and the first two are widely used in logistic regression and linear support vector machines.
Our focus is on the limited-data setting because this is where estimation errors are significant, and finding a good solution is difficult.
So we vary the number of training samples from $n=15$ to $n=200$ for each dataset. 
For each experiment, we randomly choose $n$ points as the training samples and the remainder as the test samples.
Then, we compute the average test loss for \Ourmethod and all competing methods.
We repeat this process $50$ times.
For each $n$, we report the trimmed mean of the losses, because it is  robust to the occasional outlier.
That is, we drop the five best and five worst test losses from the $50$ repetitions, and calculate the average of the remaining test losses.
We note that the mean losses have the same pattern as the trimmed means, with \Ourmethod outperforming other methods by an even wider margin.

\paragraph{Competing methods.}
The closest competitors to \Ourmethod are norm-based regularization (using $L_1$ and $L_2$ norms), and dimensionality reduction using the top few principal components (Top PCs).
For norm-based regularization, we use Python's {\texttt scikit-learn} implementations for all losses and regularizations.
We select the regularization parameter via standard cross-validation.
Note that we calculate cross-validation loss using the actual loss function which we want to minimize, and not zero-one loss as in common in practice.
Using zero-one loss gives sub-optimal results (see Appendix~\ref{app:effect_of_01}).
For Top PCs, the number of principal components is chosen by cross-validation.

We also contrast \RobustCV with two cross-validation methods.
One is the standard CV, which picks the parameter with the best average holdout loss.
The second method (\CVoneSD) considers all parameters whose loss is within one standard deviation of the best loss, and picks the parameter that achieves the most regularization~\citep{hastie01statisticallearning}.

\paragraph{Implementation details.}
For \Ourmethod\footnote{\txtred{The code is available from \url{https://faculty.mccombs.utexas.edu/deepayan.chakrabarti/mywww/software/ROLIN-2021.tgz}.}}, we run \RobustCV with $\Theta_{ratio}=5, \Theta_{slack}=0.1,$ and $\Theta_{gain}=0.05$.
We split the training data using five instances of $5$-fold cross-validation~\citep{goos_robust_2002}.
We vary $b_{max}$ from $0.01$ to $0.1$ for $n=15$.
For larger $n$, we scale the upper range with $\sqrt{n}$.
This allows more weight to be placed on the robust solution when more data is available.
We vary $\sigma_{ratio}$ from $1$ to $10$.
We also use \RobustCV to choose whether to normalize the features.
We use the above settings for all experiments except for the sensitivity analysis in Section~\ref{sec:exp:sensitivity}.

\subsection{Accuracy of \Ourmethod}
\label{sec:exp:compare}
The detailed plots of the performance of every method for each dataset and each loss function are shown in Figure~\ref{fig:all_loss} in the appendix.
In the following, we summarize our results and present our main observations.

\begin{figure}[tbp]
  \begin{center}
  \caption{
  {\em Comparison of \Ourmethod against competing methods:}
  The top row shows the number of datasets on which each method is best.
  \Ourmethod performs best for most losses and training sizes.
  The bottom row shows the ratio of the trimmed means of losses of each method against that of \Ourmethod, averaged over all $25$ datasets.
  Again, \Ourmethod works best for all loss functions and training sizes.
  We do not show $L_2$ regularization for the squared hinge and modified Huber losses, and $L_1$ regularization for the modified Huber loss, because their average loss is too large.
  }
  \label{fig:overall}
  \captionsetup[subfigure]{justification=centering}
  \begin{subfigure}{0.3\textwidth}
    \centering
    \includegraphics[width=\textwidth]{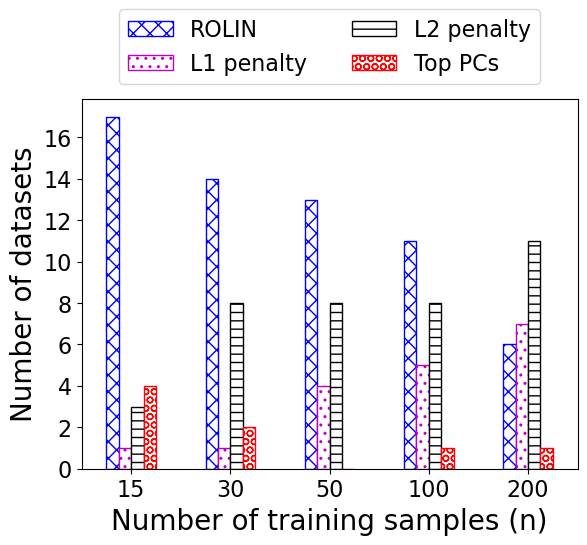}
    \caption{Logistic loss}
    \label{fig:counts_overall:logistic}
  \end{subfigure}\hspace{1em}
  \begin{subfigure}{0.3\textwidth}
    \centering
    \includegraphics[width=\textwidth]{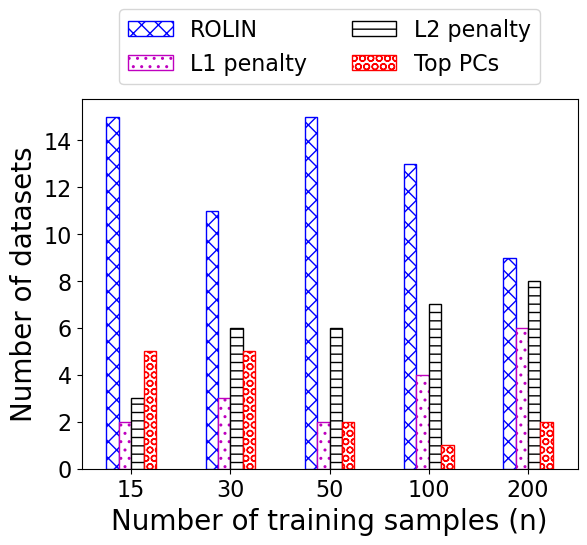}
    \caption{Squared hinge loss}
    \label{fig:counts_overall:squaredhinge}
  \end{subfigure}
  \begin{subfigure}{0.3\textwidth}
    \centering
    \includegraphics[width=\textwidth]{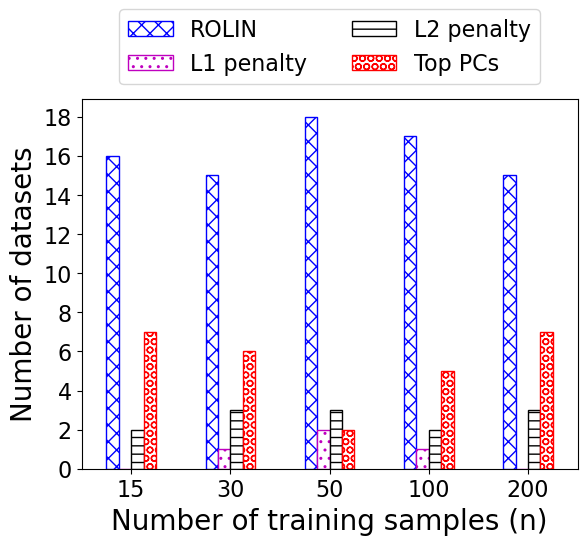}
    \caption{Modified Huber loss}
    \label{fig:counts_overall:modifiedhuber}
  \end{subfigure}
  \begin{subfigure}{0.3\textwidth}
    \centering
    \includegraphics[width=\textwidth]{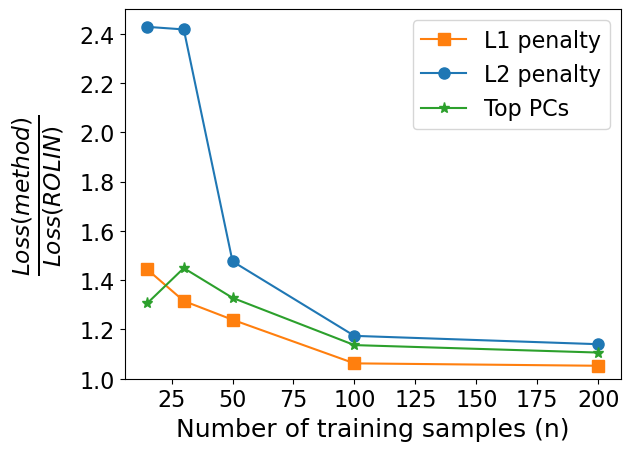}
    \caption{Logistic loss}
    \label{fig:overall:logistic}
  \end{subfigure}\hspace{1em}
  \begin{subfigure}{0.3\textwidth}
    \centering
    \includegraphics[width=\textwidth]{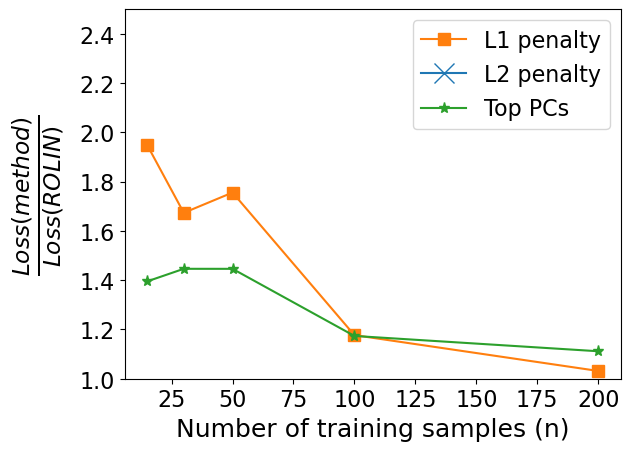}
    \caption{Squared hinge loss}
    \label{fig:overall:squaredhinge}
  \end{subfigure}
  \begin{subfigure}{0.3\textwidth}
    \centering
    \includegraphics[width=\textwidth]{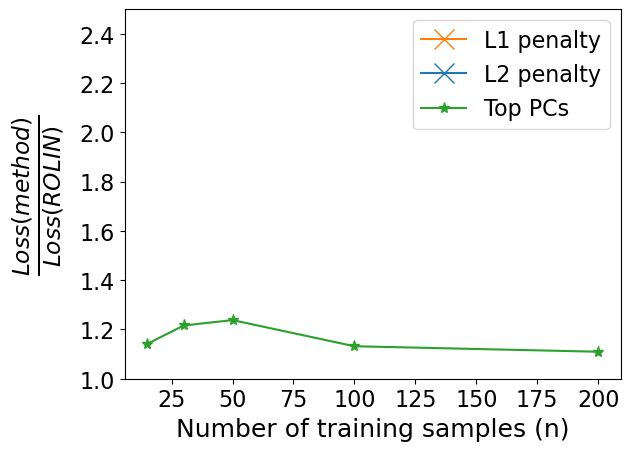}
    \caption{Modified Huber loss}
    \label{fig:overall:modifiedhuber}
  \end{subfigure}
  \end{center}
  \vspace{-2.5em}
\end{figure}
\begin{figure}[tbp]
  \begin{center}
  \caption{
  {\em Among the competitors of \Ourmethod, no one method is best:}
  We compare the three competitors of \Ourmethod against each other (ignoring \Ourmethod).
  Top PCs works well for Modified Huber loss, but for other losses, there is no one method that works best.}
  \label{fig:counts_competing_overall}
  \captionsetup[subfigure]{justification=centering}
  \begin{subfigure}{0.3\textwidth}
    \centering
    \includegraphics[width=\textwidth]{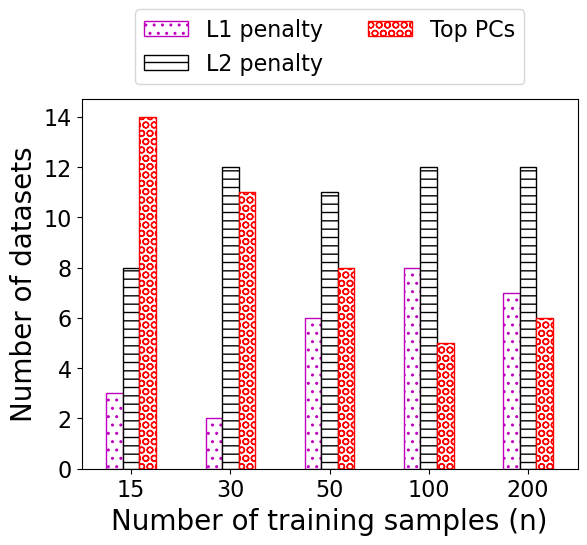}
    \caption{Logistic loss}
    \label{fig:counts_competing_overall:logistic}
  \end{subfigure}\hspace{1em}
  \begin{subfigure}{0.3\textwidth}
    \centering
    \includegraphics[width=\textwidth]{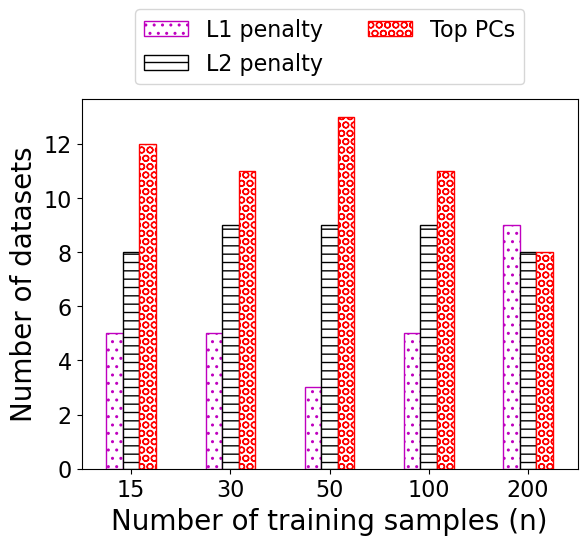}
    \caption{Squared hinge loss}
    \label{fig:counts_competing_overall:squaredhinge}
  \end{subfigure}
  \begin{subfigure}{0.3\textwidth}
    \centering
    \includegraphics[width=\textwidth]{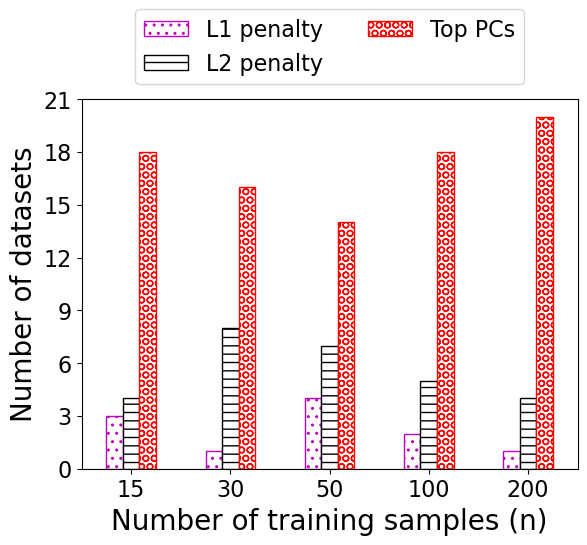}
    \caption{Modified Huber loss}
    \label{fig:counts_competing_overall:modifiedhuber}
  \end{subfigure}
  \end{center}
  \vspace{-2em}
\end{figure}
\begin{figure}[htbp]
  \begin{center}
  \caption{
  {\em \Ourmethod versus Top PCs}: \Ourmethod is consistently better.}
  \label{fig:us_vs_PCR}
  \captionsetup[subfigure]{justification=centering}
  \begin{subfigure}{0.3\textwidth}
    \centering
    \includegraphics[width=\textwidth]{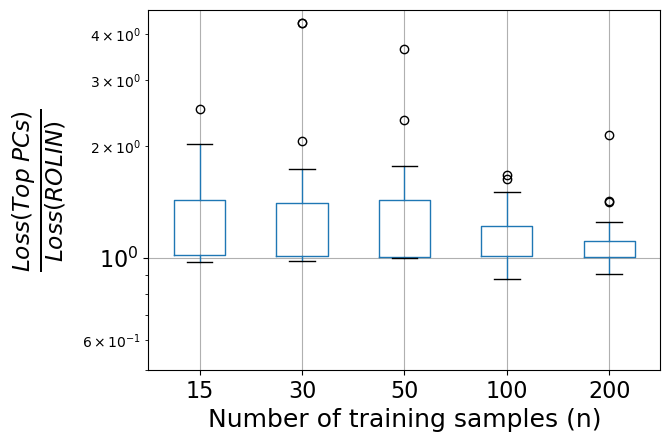}
    \caption{Logistic loss}
    \label{fig:us_PCR_boxplot:logistic}
  \end{subfigure}\hspace{1em}
  \begin{subfigure}{0.3\textwidth}
    \centering
    \includegraphics[width=\textwidth]{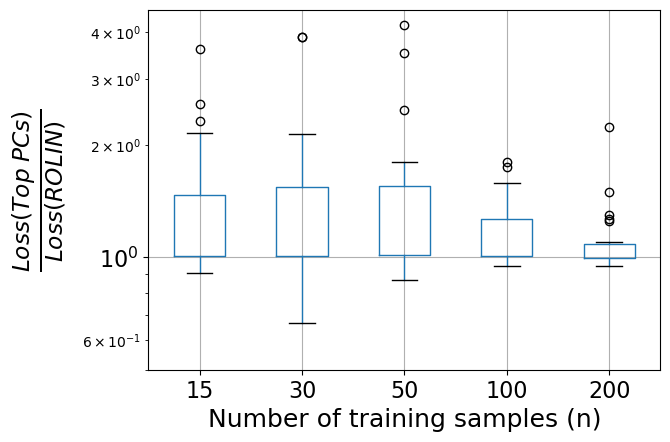}
    \caption{Squared hinge loss}
    \label{fig:us_PCR_boxplot:squaredhinge}
  \end{subfigure}
  \begin{subfigure}{0.3\textwidth}
    \centering
    \includegraphics[width=\textwidth]{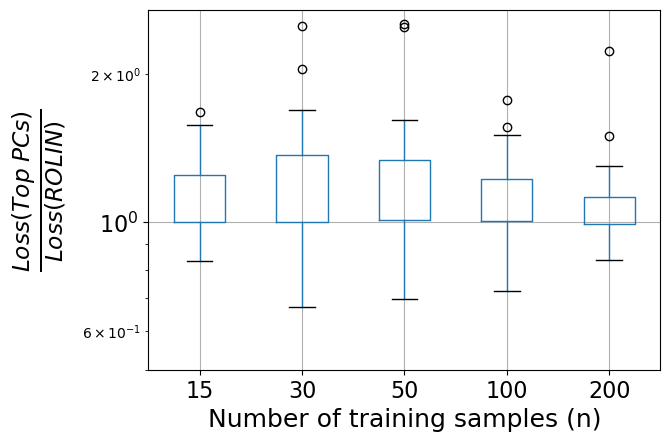}
    \caption{Modified Huber loss}
    \label{fig:us_PCR_boxplot:modifiedhuber}
  \end{subfigure}
  \end{center}
  \vspace{-2em}
\end{figure}

\paragraph{\Ourmethod outperforms the competing methods.}
Figure~\ref{fig:overall} shows the aggregate statistics comparing \Ourmethod against competing methods over all $25$ datasets.
The top panel of Figure~\ref{fig:overall} counts the number of datasets on which any given method achieves the best loss.
We see that \Ourmethod is the best in all settings except for logistic loss with $200$ training samples.
\Ourmethod is particularly dominant for small training sizes, since this is when robustness to estimation error is most needed.
For $n=15$ training samples, \Ourmethod is the best performer on at least $15$ datasets, irrespective of the loss function.
When $50$ or fewer training samples are available, \Ourmethod achieves the smallest loss on around 2x to 3x as many datasets as the next best method.
\Ourmethod also works very well for modified Huber Loss; it is best for $14$ or more datasets for any training size.

We observe the same pattern when we compare the actual value of the loss.
The bottom panel of Figure~\ref{fig:overall} shows the loss incurred by each method compared against that of \Ourmethod, averaged over all datasets.
\Ourmethod always has a better loss on average, for all loss functions and training sizes.
The greatest difference is for the smallest training size $n=15$, where the next best method is on average $14\%-40\%$ worse than \Ourmethod, depending on the loss function.
But even with $n=200$ training samples, every method is worse on average than \Ourmethod.
For the modified Huber loss, the average improvement of \Ourmethod over $L_1$ and $L_2$ regularization is too large to fit on the plot.

\paragraph{There is no clear second-best method among the competitors of \Ourmethod.}
In practice, we must choose a single method to apply to a dataset.
Figure~\ref{fig:overall} shows that while \Ourmethod is best, there is no clear second-best method.
In terms of the loss, Top PCs works well everywhere.
However, Figure~\ref{fig:overall:logistic} shows that $L_1$ regularization is better for most training sizes for logistic loss.
Further, if we consider the instances where some method outperforms \Ourmethod, that method is often $L_2$ regularization (Figures~\ref{fig:counts_overall:logistic} and~\ref{fig:counts_overall:squaredhinge}).
But the average loss for $L_2$ regularization can be much worse than the other methods (Figures~\ref{fig:overall:squaredhinge} and~\ref{fig:overall:modifiedhuber}).
Figure~\ref{fig:counts_competing_overall} compares only the competitors of \Ourmethod to each other.
Top PCs works well for the modified Huber loss, but for other losses, $L_2$ regularization is comparable or sometimes better.
Thus, among the competitors of \Ourmethod, no single method dominates.

\paragraph{Robust optimization contributes significantly to \Ourmethod's performance.}
The difference between Top PCs and \Ourmethod is that Top PCs ignores the bottom principal components, while \Ourmethod uses a robust optimization for them.
Hence, the importance of robust optimization can be gauged from the difference between these two methods\footnote{Top PCs uses standard CV while \Ourmethod uses \RobustCV. Changing Top PCs to use \RobustCV only increases the gap in performance between Top PCs and \Ourmethod.}.

The previous results show that Top PCs works reasonably well on all loss functions and training sizes.
But the consistency of Top PCs comes at a cost: it rarely outperforms \Ourmethod (top panel of Figure~\ref{fig:overall}).
Figure~\ref{fig:us_vs_PCR} shows the ratio of the loss of Top PCs against \Ourmethod on a log-scale.
For every loss function, and for any training size, \Ourmethod is better than Top PCs on at least $75\%$ of the datasets (shown by the bottom of the boxes being around one).
Further, Top PCs can be up to 4x worse than \Ourmethod.
Even with $n=200$ training samples, Top PCs can still be 2x worse.
This clearly demonstrates the need for the robust optimization step in \Ourmethod.

\paragraph{Norm-based regularization can occasionally have very large losses.}
For particular datasets and settings, \Ourmethod can be much better than both $L_1$ and $L_2$ norm-based regularizations.
It can be up to 3x better under logistic loss, and 12x better under squared hinge loss.
For modified Huber loss, no norm-based regularization yields a reasonable classifier for the {\tt Credit} and {\tt   Gas sensor} datasets (Fig.~\ref{fig:credit} and~\ref{fig:gas_sensor}).
Indeed, for several datasets, the classifiers obtained from norm-based regularization have such poor test loss that they do not appear in the plots for Figure~\ref{fig:all_loss}.

\begin{figure}[tbp]
  \begin{center}
  \caption{\small{{\em Test loss comparison on three example datasets:} \Ourmethod is compared against $L_1$ and $L_2$ regularizations for three loss functions, over a wide range of training sizes $n$.
  For each $n$, we report the trimmed mean of the test losses over $50$ repeated experiments.
	Note that the y-axis is plotted on a log-scale.}}
  \label{fig:afewdatasets}
  \captionsetup[subfigure]{justification=centering}
  \begin{subfigure}{0.32\textwidth}
    \centering
    \includegraphics[width=\textwidth]{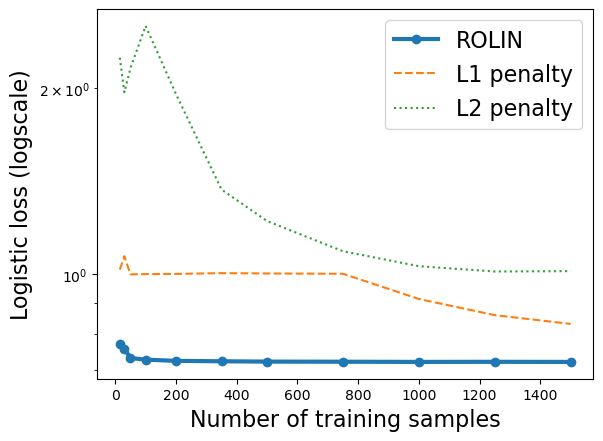}
    \caption{Logistic loss,\\ dataset {\tt Seizure} ($p=178$)}
    \label{fig:logistic_large_n}
  \end{subfigure}
  \begin{subfigure}{0.32\textwidth}
    \centering
    \includegraphics[width=\textwidth]{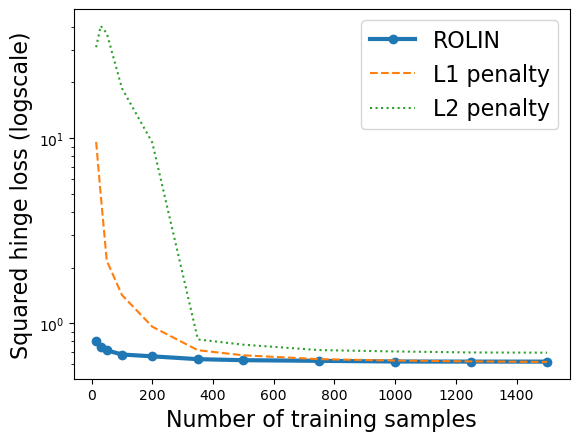}
    \caption{Squared hinge loss,\\ dataset {\tt Credit} ($p=26$)}
    \label{fig:squared_hinge_large_n}
  \end{subfigure}
  \begin{subfigure}{0.32\textwidth}
    \centering
    \includegraphics[width=\textwidth]{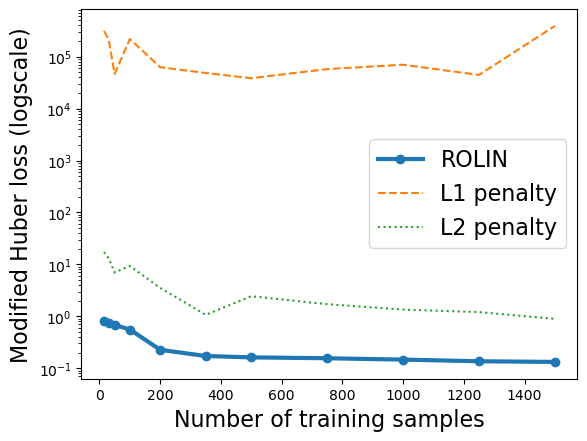}
    \caption{Modified Huber loss,\\ dataset {\tt Buzz} ($p=96$)}
    \label{fig:modified_huber_large_n}
  \end{subfigure}
  \end{center}
  \vspace{-2em}
\end{figure}

To illustrate this, Figure~\ref{fig:afewdatasets} compares \Ourmethod against norm-based regularization for three specific datasets.
Note that the y-axis is on a log scale, and we report trimmed means which remove outliers.
Plot~\subref{fig:logistic_large_n} shows an instance where \Ourmethod with $n=15$ training samples is better than both norm-based methods with $n=1500$ samples.
Plot~\subref{fig:modified_huber_large_n} shows a similar situation.
In plot~\subref{fig:squared_hinge_large_n}, the losses for norm-based regularization become much worse when training size is reduced.
This cannot be due to occasional outliers, because the trimmed mean ignores the worst five test losses.
Further, $L_1$-regularization in plot~\subref{fig:modified_huber_large_n} is not close to convergence even with $n=1500$.
These examples highlight the perils of choosing the ``wrong'' norm.
\Ourmethod sidesteps this issue entirely.

\txtred{
  Finally, we note that \Ourmethod performs as well or better than competing methods for zero-one loss (or, misclassification rate).
  Since our focus is on convex losses, we defer these results to  Appendix~\ref{app:effect_of_01}.
}

\begin{figure}[tbp]
  \begin{center}
  \caption{
    \small{
  {\em Importance of \RobustCV:}
We plot the loss when \Ourmethod is run with CV and \CVoneSD, versus \RobustCV.
\RobustCV is much better for small training sizes.}}
  \label{fig:standardCV}
  \begin{subfigure}{0.3\textwidth}
    \centering
    \includegraphics[width=\textwidth]{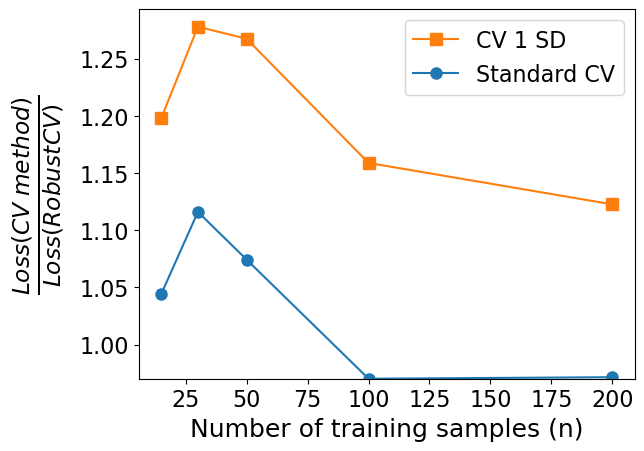}
    \caption{Logistic loss}
    \label{fig:CV:logistic}
  \end{subfigure}\hspace{1em}
  \begin{subfigure}{0.3\textwidth}
    \centering
    \includegraphics[width=\textwidth]{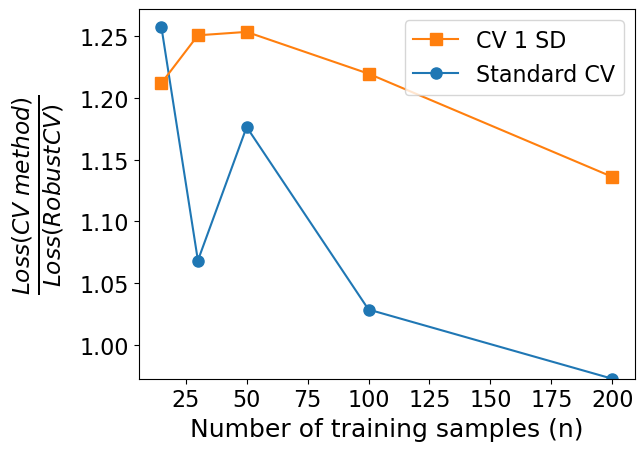}
    \caption{Squared hinge loss}
    \label{fig:CV:squaredHinge}
  \end{subfigure}\hspace{1em}
  \begin{subfigure}{0.3\textwidth}
    \centering
    \includegraphics[width=\textwidth]{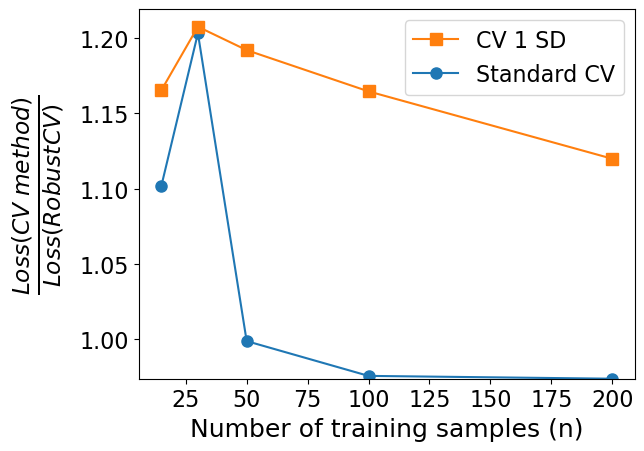}
    \caption{Modified Huber loss}
    \label{fig:CV:modifiedHuber}
  \end{subfigure}
  \end{center}
  \vspace{-2em}
\end{figure}
\begin{figure}[tbp]
  \begin{center}
  \caption{\small{{\em Sensitivity to the parameters of \RobustCV:}
We plot the relative difference in trimmed means for logistic loss when the parameters $(\Theta_{ratio}, \Theta_{slack}, \Theta_{gain})$ are varied from their default values of $(5, 0.1, 0.05)$. 
Positive values imply larger losses.
\Ourmethod is seen to be robust to a wide range of parameter choices.}}
  \label{fig:sensitivity}
  \begin{subfigure}{0.31\textwidth}
    \centering
    \includegraphics[width=\textwidth]{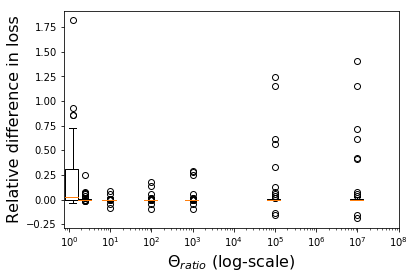}
    \caption{Varying $\Theta_{ratio}$}
    \label{fig:theta_ratio}
  \end{subfigure}\hspace{1em}
  \begin{subfigure}{0.31\textwidth}
    \centering
    \includegraphics[width=\textwidth]{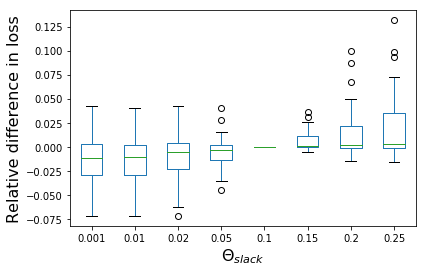}
    \caption{Varying $\Theta_{slack}$}
    \label{fig:theta_slack}
  \end{subfigure}\hspace{1em}
  \begin{subfigure}{0.31\textwidth}
    \centering
    \includegraphics[width=\textwidth]{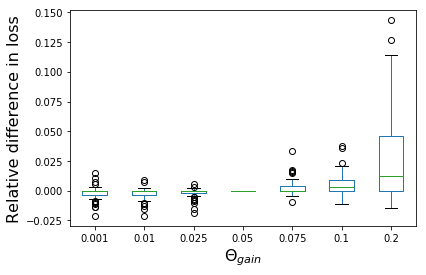}
    \caption{Varying $\Theta_{gain}$}
    \label{fig:theta_gain}
  \end{subfigure}
  \end{center}
  \vspace{-2em}
\end{figure}

\subsection{Importance of \RobustCV}
\label{sec:exp:robustcv}
To examine the influence of \RobustCV, we run \Ourmethod with standard cross-validation (CV) and another common variant (\CVoneSD).
For each dataset and loss function, we calculate the trimmed mean of test loss of \Ourmethod with CV and \CVoneSD, for $n=15$ to $n=200$.
We compare these against the trimmed means using \RobustCV.
Figure~\ref{fig:standardCV} show that CV is better than \CVoneSD for all losses.
Between CV and \RobustCV, \RobustCV outperforms for small training sizes.
For $n\leq 30$, CV is $5\%-25\%$ worse on average than \RobustCV, depending on the loss function.
The differences mostly disappear when more training samples are available.



These results show the usefulness of robustness in cross-validation for small sample sizes.
In such scenarios, an overconfident classifier may correctly classify all but a few points, and only these few points provide any warning about the unsuitability of the classifier.
\RobustCV is designed to look for these warning signals and hence can avoid overconfident classifiers.
Standard CV averages over all holdout sets, and this attenuates or even hides the warning signs.

\vspace{-1em}
\subsection{Sensitivity Analysis}
\label{sec:exp:sensitivity}
Recall that \RobustCV requires three parameters.
The first is $\Theta_{ratio}$, which is the threshold ratio of holdout to training loss above which we distrust the average holdout loss.
The second is $\Theta_{slack}$, which is the importance we assign to the maximum holdout loss versus the average holdout loss.
The third is $\Theta_{gain}$, which characterizes our preference for solutions constructed only from the top principal components.
Now, we vary these parameters one at a time from their default values and report results on ten datasets.

Figure~\ref{fig:sensitivity} shows the relative increase in the trimmed means under logistic loss for different values of these parameters.
Plot~\subref{fig:theta_ratio} shows that for $\Theta_{ratio}$, any value in the range $\Theta_{ratio}\in[2.5, 100]$ works well (the default is $5$).
Larger values of $\Theta_{ratio}$ mean that we ignore instances where training loss is much smaller than holdout loss, which is a clear sign of overfitting.
Smaller values mean that we always use the maximum holdout loss instead of the average holdout loss.
Always focusing on maximum loss is too conservative, so it performs poorly for our expected test loss objective. 

Plot~\subref{fig:theta_slack} shows that any choice of $\Theta_{slack}\leq 0.15$ yields similar results  (the default is $0.1$).
Losses become worse for larger values of $\Theta_{slack}$.
A large $\Theta_{slack}$ means that we downplay the average holdout loss and focus on the maximum holdout loss.
Like a small $\Theta_{ratio}$, this is too conservative and does not work for the same reason.
 
Plot~\subref{fig:theta_gain} shows that any $\Theta_{gain}\leq 0.05$ yields good results (the default is $0.05$).
Higher values imply a preference for solutions based on only the top few principal components, ignoring the robust solution from the remaining principal components.
When $\Theta_{gain}\to\infty$, we get Top PCs.
We see that high $\Theta_{gain}$ leads to a significant increase in the test loss, showing the importance of the robust component of \Ourmethod.

Extreme values for any of these parameters correspond to either standard cross-validation or very conservative choices.
The former is bad for small $n$, while the latter performs poorly for large $n$.
But for a broad range of parameters, \RobustCV achieves good results.

%% file: related.tex
\section{Prior work}
\label{sec:related}
A common approach to deal with limited data is regularization.
Here, we add to the desired objective an extra term that penalizes large feature weights.
This term is typically some $L_q$ norm of the feature weight vector, with $q=1$ and $q=2$ being common choices:
\begin{align}
\min_{\bb} \mathbb{P}_n \ell(y\cdot \gbx) + \lambda\cdot\|\bbw\|_q.
\label{eq:reg}
\end{align}
There are several competing justifications of the regularization term in Eq.~\ref{eq:reg}.
Regularization can emerge from a prior, or as the solution of a robust optimization, or as a way to bound estimation errors.
Next, we discuss these, and contrast them with \Ourmethod.

\paragraph{Prior.}
We can cast regularization as a prior on the parameter vector $\bb$.
Then, the solution of Eq.~\ref{eq:reg} is the {\em maximum a posteriori} (MAP) estimate of $\bb$.
For example, a zero-mean spherical Gaussian prior for \bbw gives $L_2$ regularization, while a Laplace prior yields $L_1$ regularization.
But one may construct a prior for any $L_q$-norm, or any Mahalanobis distance metric.
Choosing the best prior for a dataset is difficult, but it matters a lot, as we showed in Section~\ref{sec:exp}.
\Ourmethod does not assume a prior, so it sidesteps this difficulty entirely.

Priors are also useful for dealing with corrupted data~\citep{kordzakhia_robust_2001, feng_robust_2014, tibshirani_robust_2014}.
Further, $L_1$ priors induce sparsity in the solution, which makes the model easier to interpret~\citep{tibshirani_regression_1996}.
We do not consider data corruption or interpretability in this paper.

\paragraph{Robust optimization.}
Many optimization problems have parameters or constraints that must be learned from data. 
Robust optimization methods protect against corrupted data, outliers, and incorrect assumptions~\citep{ben2009robust}.
These methods first construct uncertainty sets that reflect the ambiguity in the data.
Then, they optimize a worst-case objective over the uncertainty set. 
For some uncertainty sets, this worst-case objective matches norm-based regularization.

Robust optimization methods typically fall into two groups.
Methods in the first group assume that the training samples are perturbed.
The perturbation could be because of uncertain or missing data~\citep{trafalis_robust_2006, gao_huang_robust_2012, wang_survey_2014, tzelepis_linear_2018}, adversarial opponents~\citep{globerson06nightmare}, or different training and test distributions~\citep{bi_support_2004}.
Robustness to perturbations is also equivalent to robustness under chance constraints~\citep{bhattacharyya_robust_2004, shivaswamy_second_2006}.
To achieve robustness, we assume that the ``true'' data fall inside an uncertainty set constructed from the ``perturbed'' data.
Standard uncertainty sets impose a bound on some norm of the perturbation.
Choosing a particular norm gives a corresponding norm-based regularization~\citep{el1997robust, xu2009robust, xu_robustness_2009}.


The second group of robust optimization methods constructs uncertainty sets of probability distributions.
They assume that the true distribution of $(y, \bx)$ lies in this uncertainty set and optimize for the worst-case distribution in this set.
\citet{delage_distributionally_2010, goh_distributionally_2010, wiesemann_distributionally_2014} consider distributions with appropriately bounded moments.
Others choose distributions within a bounded distance from the empirical distribution.
The distance can be the Prohorov metric~\citep{erdogan_ambiguous_2006}, KL-divergence~\citep{jiang_datadriven_2016}, or Wasserstein distance~\citep{wozabal_framework_2012, shafi15distributionally, shafi17regularization, mohajerin_esfahani_datadriven_2018}. 
For Wasserstein distance, the user must also choose a distance metric in feature space.
Choosing a distance metric based on some norm yields a regularization using that norm.

Thus, both types of robust optimization approaches rely on the user to choose a distance metric or a norm.
This choice determines the form of the regularization term.
The ``best'' choice for a dataset is unclear.
Further, robust optimization emphasizes worst-case performance.
This can make robust algorithms too conservative for our average-loss objective.
In contrast, \Ourmethod does not require any user inputs.
Also, \Ourmethod restricts robust optimization to just the bottom principal components, which are much noisier than the top components.
This protects \Ourmethod from becoming too conservative.

\paragraph{Estimation error bounds.}
Regularization also ensures that the training loss is close to the expected loss.
Let $\Delta_\bb := E\ell(y\cdot \gbx) - \mathbb{P}_n \ell(y\cdot \gbx)$ be the difference between the expected and training losses.
For large training sizes $n$, this is small for any \bb.
But, for small $n$, $\Delta_\bb$ can be much greater than zero for some values of \bb.
However, if $\bb$ has a small norm, $\Delta_\bb$ can be upper-bounded.
For example, under zero-one loss, if $\|\bb\|_1\leq 1/\rho$ (bounded $L_1$ norm), then $\Delta_\bb$ decays with $\rho$ and the square-root of $n$~\citep[see][]{mohri12foundations}.
Similar results hold when \bb has a small $L_2$ norm.
Now, regularization biases the objective of Eq.~\ref{eq:reg} towards a \bb with a small norm.
This ensures that the expected loss of the solution is comparable to its training loss.
Hence, regularization avoids overfitting.

Still, bounding $\Delta_\bb$ is not enough.
Our aim is a \bb with small expected loss, not a small $\Delta_\bb$.
Further, such bounds may hold for many norms or Mahalanobis distances.
Choosing the best norm for a dataset is difficult.
For small $n$, the bounds on $\Delta_\bb$ can be loose\footnote{For instance, if the bound needs to hold with probability greater than $0.95$, then the error term in the bound is at least $0.13$ for $n\leq 100$~\citep[see][Thms.~13.3 and 13.4]{mohri12foundations}. For comparison, the zero-one loss of even a baseline classifier is at most $0.5$.}.
So, we cannot just pick the norm with the best bound.
Finally, while a \bb with a small norm may have a small $\Delta_\bb$, the converse need not be true.
There may be other solutions that have a small $\Delta_\bb$ and also a low loss.


All the above justifications for regularization need the user to choose a norm, or a prior, or a distance metric.
The right choice depends on the dataset, the training size, and the loss function.
This choice is challenging but also crucial because the wrong choice can significantly hurt performance.
In contrast, \Ourmethod needs no user input and does not force the solution to have a small norm.
This suggests that explicit norm-based regularization of the form of Eq.~\ref{eq:reg} is unnecessary.

%% file: conc.tex
\section{Conclusions}
\label{sec:conc}

Our goal is to build a linear classifier with two properties.
First, it should optimize for general loss functions, instead of the usual zero-one loss.
This can be interpreted as accurately predicting class probabilities and not just the binary class labels.
Second, its accuracy should gracefully degrade with smaller training sample sizes.
The usual approach is to do dimensionality reduction via principal components, or to add to the loss function a regularization term based on a norm chosen by the user.
But dimensionality reduction loses data, while regularization is sensitive to the choice of norm.
Our proposed method, called \Ourmethod, overcomes these flaws.
Unlike dimensionality reduction, it does not ignore the bottom principal components.
Unlike regularization, \Ourmethod is entirely automatic and needs no user input.
Further, it works well with many loss functions.

\Ourmethod first projects the data on to its top principal components and minimizes training loss on the projected data.
The resulting classifier does not overfit because the top principal components are stable.
But this classifier ignores the subspace orthogonal to the top principal components.
We cannot minimize training loss in this subspace, because estimates of loss are unreliable.
So \Ourmethod constructs a robust classifier here.
Finally, \Ourmethod combines the two classifiers to get the benefits of both.

To select the parameters of \Ourmethod, we develop a new robust cross-validation algorithm called \RobustCV.
This checks for several warning signs of overfitting missed by standard cross-validation.
\RobustCV helps \Ourmethod work well even with small training sizes.

Experiments on $25$ real-world datasets and three loss functions show that \Ourmethod outperforms existing state of the art methods.
\Ourmethod does particularly well for small training sizes.
For $n=15$ training samples, \Ourmethod has $14\%-40\%$ lower loss on average than the next-best competitor, under all problem settings.
When $50$ or fewer training samples are available, \Ourmethod achieves the smallest loss on around 2x to 3x as many datasets as the next best method.
For the modified Huber loss, \Ourmethod dominates other methods for all training sizes
Further, among the competitors of \Ourmethod, no single method is best.
Norm-based regularization is close to \Ourmethod for logistic regression, but is not comparable for modified Huber loss.
On some datasets, \Ourmethod achieves with $n=15$ samples an accuracy that regularization fails to reach with $n=1500$ samples.
Dimensionality reduction via the top principal components rarely outperforms \Ourmethod, especially for logistic loss.
Finally, the best norm for regularization depends on the dataset, training size, and loss function.
So, for a new problem setting, picking the right norm is difficult.
In contrast, \Ourmethod works well for all datasets and settings.

There are several ways to extend \Ourmethod.
We can try to use \Ourmethod for non-linear classification via the kernel trick.
Here, each test point \bx is classified based on a linear combination of $K(\bx, \bx_i)$ where $\bx_i$ is a training point and $K(.,.)$ is a kernel function.
This suggests that we can use \Ourmethod  on the kernel matrix instead of the features matrix. 
\txtred{We can also use \Ourmethod for multiclass classification via one-versus-the-rest binary classification.}
Finally, we note that \Ourmethod does not handle outliers or different training and test distributions.
The top principal components of the training and test distributions may not be similar in this setting.
One possibility is to project \Ourmethod's solution on to the set of small-norm solutions, which may be more robust under outliers.

%% file: appendix.tex
\section*{Appendices}
\section{Proofs}
\label{app:theorem}



\begin{proof}[Theorem~\ref{thm:main}]
Consider the case of a loss function $\ell(.)$ that satisfies the properties stated in the theorem.
Choose any feasible ${\bm b}\in\Sot$, and let $\tilde{\bm b} = \Vot^T {\bm b}$.
Choose any $B_{22}\in\mathcal{U}$.
This fixes the second moment of $P_{\Sot}\bz$ as $\Vot \Sigma_{B_{22}} \Vot^T$, where
\begin{align*}
\Sigma_{B_{22}} &=: \begin{bmatrix} B_{11} & 0 \\ 0 & B_{22} \end{bmatrix},
\end{align*}
with $B_{11} = \mathbb{P}_n \left[ \left(\Vone^T\bz\right) \left(\Vone^T\bz\right)^T \right]$.
Also, from Eq.~\ref{eq:expect2}, the first moment is fixed at $(1/n) \sum_i P_{\Sot}\bz_i = \Vot {\bm \mu}$.
Now, the maximum entropy distribution with given first and second moments is the Gaussian distribution with those moments:
\begin{align*}
q(.) &= \mathcal{N}\left( \Vot{\bm \mu}, \Vot\left(\Sigma_{B_{22}} - {\bm \mu}{\bm \mu}^T\right)\Vot^T \right).
\end{align*}
Since ${\bm r}\sim q(.)$, for any training sample $i$, we have
\begin{align}
\beta_0\cdot y_i + \bbSz^T \left(P_{\Szero} \bz_i\right) + {\bm b}^T {\bm r} &\sim \mathcal{N}(m_i, s^2),\\
\text{where } m_i &= \beta_0\cdot y_i + \bbSz^T \left(P_{\Szero} \bz_i\right) + \tilde{\bm b}^T {\bm \mu}\\
s &= \tilde{\bm b}^T \left(\Sigma_{B_{22}} - {\bm \mu}{\bm \mu}^T\right) \tilde{\bm b}.
\label{eq:mi_s}
\end{align}
Hence, the expected loss under $q(.)$ is
\begin{align}
\frac{1}{n} \sum_{i=1}^n E_{{\bm r} \sim q(.)} \left[ \ell\left(\beta_0\cdot y_i + \bbSz^T \left(P_{\Szero} \bz_i\right) + {\bm b}^T {\bm r}\right)\right]
  &= \frac{1}{n} \sum_{i=1}^n E_{w\sim\mathcal{N}(m_i, s^2)}[\ell(w)]\nonumber\\
  &= \frac{1}{n} \sum_{i=1}^n E_{w\sim\mathcal{N}(0, 1)}[\ell(m_i + w\cdot s)].\label{eq:proof_eloss}
\end{align}

Taking the partial derivative with respect to $s$, we have
\begin{align*}
\frac{\partial}{\partial s} E_{w\sim\mathcal{N}(0, 1)}[\ell(m_i + w\cdot s)] 
  &= E[w\cdot\ell'(m_i + w\cdot s)]\\
  &= \int_0^\infty w\left[\ell'(m_i + w\cdot s) - \ell'(m_i - w\cdot s)\right]\cdot \phi(w) dw\\
  &> 0,
\end{align*}
where the interchange of differentiation and integration in the first equality follows from the dominated convergence theorem since $|\ell'(.)|$ has finite expectation, and the last inequality is because $\ell'(.)$ is monotonically non-decreasing (due to convexity) and not zero everywhere (because $|\ell'(.)|$ has non-zero expectation).
Thus, for any ${\bm b}$, the worst case expected loss is achieved when $s=\tilde{\bm b}^T (\Sigma_{B_{22}} - {\bm \mu}{\bm \mu}^T) \tilde{\bm b}$ is maximized.
From the uncertainty set of Eq.~\ref{eq:B22}, the maximum is achieved at $B_{22} = \sigma_{bound}\cdot I$.
Hence, the worst-case is achieved with $\Sigma_{B_{22}} = \Sigma$, and $s = \tilde{\bm b}^T \left( \Sigma - {\bm \mu}{\bm \mu}^T \right) \tilde{\bm b}$, where $\Sigma$ is defined in the theorem statement.

Next, taking the partial derivative with respect to $m_i$, we have
\begin{align*}
\frac{\partial}{\partial m_i} E_{w\sim\mathcal{N}(0, 1)}[\ell(m_i + w\cdot s)] 
  &= E_{w\sim\mathcal{N}(0, 1)}[\ell'(m_i + w\cdot s)] < 0,
\end{align*}
where $\ell'(.)\leq 0$ because $\ell(.)$ is monotonically non-increasing, and $\ell'(x)<0$ for some $x$ because $|\ell'(.)|$ has non-zero expectation.
Now, observe that each $m_i$ (for $i=1\ldots n$) increases with $\tilde{\bm b}^T{\bm \mu}$, 
Hence, if Eq.~\ref{eq:bbSot} achieves its optimal at $s=s^\star$, then the optimal solution \bbSot is of the form $\bbSot=\Vot\tilde{\bm b}$, where $\tilde{\bm b}$ solves
\begin{align*}
\text{maximize }& \tilde{\bm b}^T {\bm \mu}\\
\text{subject to }& \tilde{\bm b}^T \left( \Sigma - {\bm \mu}{\bm \mu}^T \right) \tilde{\bm b} = s^\star.
\end{align*}
It is easily shown that the solution must be of the form
\begin{align}
\bbSot = c\cdot \Vot\left(\Sigma - {\bm \mu}{\bm \mu}^T\right)^{-1} {\bm \mu},
\label{eq:bbSotNearFinal}
\end{align}
for some scalar $c$ that depends on $s^\star$.
Finally, by the Sherman-Morrison formula,
\begin{align*}
\left(\Sigma - {\bm \mu}{\bm \mu}^T\right)^{-1} {\bm \mu} &= \frac{1}{1 - {\bm \mu}^T \Sigma^{-1} {\bm \mu}}\cdot \Sigma^{-1}\mu.
\end{align*}
Combining this with Eq.~\ref{eq:bbSotNearFinal} gives the statement of the theorem.

Now, consider a loss function $\ell(.)$ that is the limit of a sequence of loss functions $\ell^{(m)}(.)$ such that $\lim_{m\to\infty}\sup_{x\in\mathbb{R}} |\ell^{(m)}(x) - \ell(x)| = 0$.
Then, for any $\epsilon>0$, there exists an $N$ such that for all $m>N$, $\sup_{x\in\mathbb{R}} |\ell^{(m)}(x) - \ell(x)| < \epsilon$.
Define $h({\bm b}) = 1/n\cdot \sum_i E[\ell(m_i + w\cdot s)]$, where $m_i$ and $s$ are functions of ${\bm b}$; define $h^{(m)}({\bm b})$ similarly.
Then, for any ${\bm b}$ and for large enough $m$,
\begin{align}
\left| h^{(m)}({\bm b}) - h({\bm b})\right| < \epsilon.
\label{eq:proof_limit1}
\end{align}
There is a sequence of minimizers ${\bm b}^{(m)}$ under $\ell^{(m)}$ such that ${\bm b}^{(m)} = c^{(m)}\cdot \Sigma^{-1}{\bm \mu}$.
Hence,
\begin{align*}
h^{(m)}\left({\bm b}^{(m)}\right) &\leq h^{(m)}\left(\bbSot\right) \leq h^{(m)}\left({\bm b}^{(m)}\right) + 2\epsilon,
\end{align*}
where the first statement follows from the optimality of ${\bm b}^{(m)}$ under $\ell^{(m)}$, and the second from two applications of Eq.~\ref{eq:proof_limit1}.
So, the minimizers under $\ell^{(m)}$ provide arbitrarily good solutions under $\ell$, for $m$ large enough.
\end{proof}

\begin{proof}[Corollary~\ref{cor:losses}]
The logistic loss $\ell(x) = \log(1+\exp(-x))$ is clearly non-negative, monotonically decreasing, and convex.
Also, $|\ell'(x)| = \exp(-x) / (1+\exp(-x))$, so $0 < |\ell'(x)| < 1$, so it has finite non-zero expectation under the Gaussian.
Hence, the conditions of Theorem~\ref{thm:main} are satisfied.

The squared hinge loss $\ell(x) = (\max(0, 1-x))^2$ is non-negative, monotonically decreasing, and convex.
Also, $|\ell'(x)| = 2\cdot\max(0, 1-x) \leq 2(1+|x|)$, so $0 < E|\ell'(x)| \leq 2 + 2\sqrt{2/\pi}$.
Hence, Theorem~\ref{thm:main} applies.

The hinge loss $\ell(x) = \max(0, 1-x)$ is the limit of a sequence of functions $\ell^{(m)}(.)$ indexed by a monotonically decreasing sequence $\alpha_m\to 0$:
\begin{align*}
\ell^{(m)}(x) = \left\{ 
  \begin{array}{cl}
  \max(0, 1-x) & \text{if } x\leq 1-\alpha_m\\
  \frac{\left( x - (1+\alpha_m) \right)^2}{4\alpha_m} & 1-\alpha_m < x \leq 1 + \alpha_m\\
  0 & x > 1+\alpha_m
  \end{array}
  \right.
\end{align*}
Each $\ell^{(m)}(.)$ is non-negative, monotonically decreasing, and convex.
Also, $|\ell^{(m)'}(x)| \leq 1$ with equality for $x\leq 1-\alpha_m$, so it has finite non-zero expectation.
Again, Theorem~\ref{thm:main} applies.

The modified Huber loss is similar to the functions $\ell^{(m)}(.)$ above, for which the conditions of the Theorem are satisfied.
\end{proof}

\begin{proof}[Theorem~\ref{thm:ridge}]
Let $\tilde{Z}$ be a matrix with $\tilde{\bz}_i := P_\Sot\bz_i$ in row $i$.
Then, using $\Sigma_{smooth}$ in Theorem~\ref{thm:main},
\begin{align}
\bbSot 
  &\propto \Vot \Sigma_{smooth}^{-1} \Vot^T \left(\sum_i \bz_i/n\right)\nonumber\\
  &= \left(\Vot \Sigma_{smooth} \Vot^T\right)^+ \left(\sum_i P_\Sot\bz_i/n\right)\nonumber\\ 
  &= \left(\Vot D_{\Sot}^2 \Vot^T + n\sigma_{bound}\cdot \Vot\Vot^T\right)^+ \left(\tilde{Z}^T\bone\right)\nonumber\\
  &= \left(\tilde{Z}^T\tilde{Z} + n\sigma_{bound}\cdot \Vot\Vot^T\right)^+ \left(\tilde{Z}^T\bone\right)\nonumber\\ 
  &= \left(\tilde{Z}^T\tilde{Z} + n\sigma_{bound}\cdot I\right)^{-1} \left(\tilde{Z}^T\bone\right),\label{eq:ridge} 
\end{align}
where $M^+$ is the Moore-Penrose pseudoinverse of matrix $M$, and \bone is a vector of all ones.
The first equality follows from the observation that $\Sigma_{smooth}$ is diagonal, and the fact that $P_\Szero\bz$ is in the kernel of the pseudoinverse.
The change from $\Vot\Vot^T$ to the identity matrix in the last step makes the matrix invertible but does not change the solution, since the rows of $\tilde{Z}$ lie in the subspace \Sot.
Eq.~\ref{eq:ridge} is the solution of the following ridge regression problem:
\begin{align*}
\min_{\bm b} \sum_{i=1}^n \left(1-{\bm b}^T\tilde{\bz}_i\right)^2 + n\sigma_{bound}\cdot \|{\bm b}\|^2
  &= \min_{\bm b} \sum_{i=1}^n \left(y_i-{\bm b}^T\tilde{\bx}_i\right)^2 + n\sigma_{bound}\cdot \|{\bm b}\|^2,
\end{align*}
where $\tilde{\bx}_i = P_\Sot\bx_i$, and we use the fact that $y_i\in\{-1,1\}$.
\end{proof}

\begin{figure}[tbp]
  \begin{center}
	\caption{\small{\txtred{{\em Zero-one loss:} 
We compare the misclassification loss of \Ourmethod against other methods.
Positive values imply that \Ourmethod is more accurate than the competing method.
We pick the feature weights that minimize logistic loss but choose the hyperparameters by cross-validation to minimize zero-one loss.
\Ourmethod outperforms $L_1$ regularization and Top-PCs, and is comparable to $L_2$ regularization.
}}}
	\label{fig:optimize_01}
	\includegraphics[width=\textwidth]{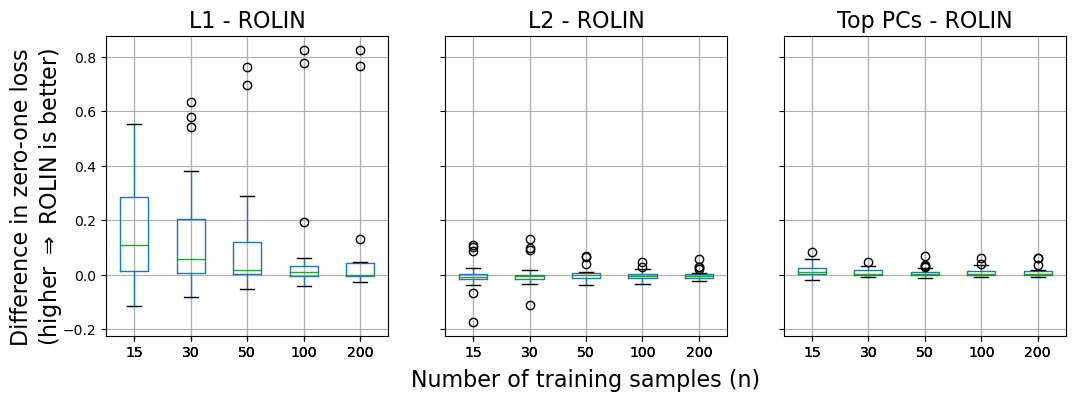}
	\end{center}
  \vspace{-2em}
\end{figure}
\begin{figure}[htbp]
  \begin{center}
  \caption{\small{{\em Relative difference in loss if cross-validation is run with zero-one loss instead of logistic loss:}
	Optimizing for zero-one loss in cross-validation leads to worse logistic loss on the test set.
	In other words, the predicted class probabilities are inaccurate under cross-validation with zero-one loss.}}
	\label{fig:effect_01}
  \begin{subfigure}{0.4\textwidth}
    \centering
    \includegraphics[width=\textwidth]{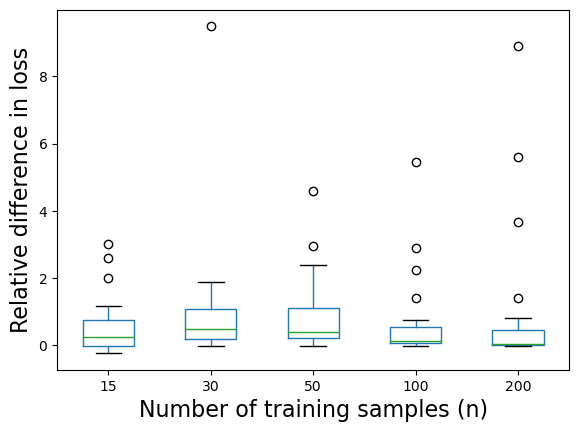}
    \caption{$L_1$ regularization}
    \label{fig:effect_01_L1}
  \end{subfigure}\hspace{1em}
  \begin{subfigure}{0.4\textwidth}
    \centering
    \includegraphics[width=\textwidth]{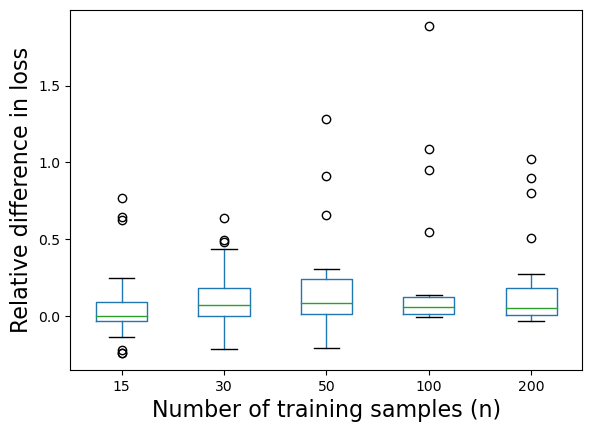}
    \caption{$L_2$ regularization}
    \label{fig:effect_01_L2}
  \end{subfigure}
  \end{center}
\end{figure}

\section{Results for Zero-One Loss}
\label{app:effect_of_01}

\txtred{
In many applications, we only care about zero-one (misclassification) loss.
For such tasks, we pick the hyperparameters (e.g., the regularization $\lambda$ in Eq.~\ref{eq:reg}) that minimize zero-one loss in cross-validation.
After fixing these hyperparameters, we optimize the feature weights using a convex loss such as logistic loss.
Thus, the parameter-fitting step uses two losses.
In contrast, our previous experiments used a single convex loss everywhere.
We now explore the effects of this ``double-loss'' optimization.
}

\txtred{
Figure~\ref{fig:optimize_01} shows the misclassification loss of various methods after ``double-loss'' parameter-fitting.
\Ourmethod outperforms $L_1$ regularization and Top-PCs, and is similar to $L_2$ regularization.
Thus, even though \Ourmethod is aimed at convex losses, it is useful even for zero-one loss.
}


\txtred{
However, the double-loss optimization improves zero-one loss only at the cost of worse values for the logistic loss.
Figure~\ref{fig:effect_01} shows the relative increase in the logistic loss on test samples when we move from the single-loss to the double-loss parameter-fitting.
The test logistic loss increases for both $L_1$ and $L_2$ regularization for all training sizes.
The test logistic loss can be up to $10x$ larger for $L_1$ and up to $2x$ larger for $L_2$ regularization.
Hence, the confidence scores (or equivalently, the class probabilities) learned by the logistic regression classifier can be inaccurate if we use zero-one loss to choose the hyperparameters.
}

\renewcommand\thesubfigure{(\arabic{subfigure})}
\input{loss_figures}
\renewcommand\thesubfigure{(\alph{subfigure})}

%% file: loss_figures.tex
\begin{figure}[!b]
  \caption{\small{Logistic, squared hinge, and modified Huber losses for 25 datasets under varying training sizes. Each plot compares the trimmed mean loss of \Ourmethod against $L_1$ and $L_2$ regularization. Regularized losses are too large to fit in some plots, such as plots~\subref{fig:buzz} and~\subref{fig:seizure}.}}
  \label{fig:all_loss}
    \begin{subfigure}{\textwidth}
      \centering
      \includegraphics[width=0.28\linewidth]{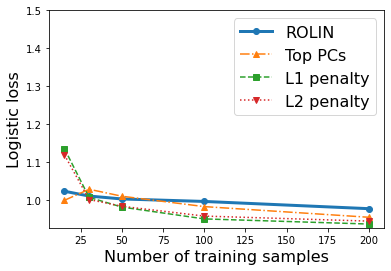}
      \includegraphics[width=0.28\linewidth]{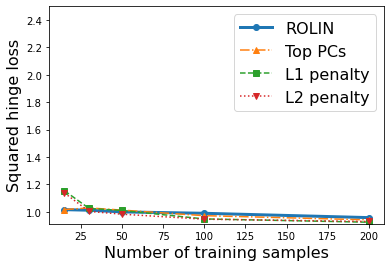}
      \includegraphics[width=0.28\linewidth]{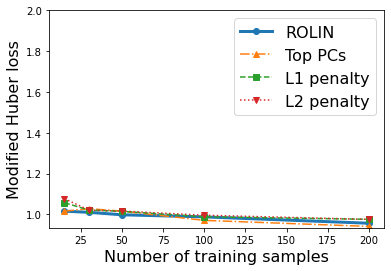}
      \caption{{\tt Yeast} (p=8)}
      \label{fig:yeast}
    \end{subfigure}
    \begin{subfigure}{\textwidth}
      \centering
      \includegraphics[width=0.28\linewidth]{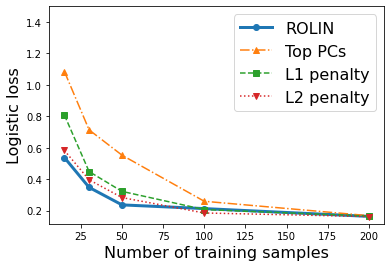}
      \includegraphics[width=0.28\linewidth]{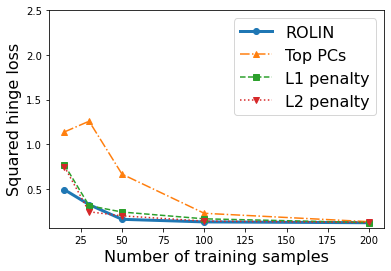}
      \includegraphics[width=0.28\linewidth]{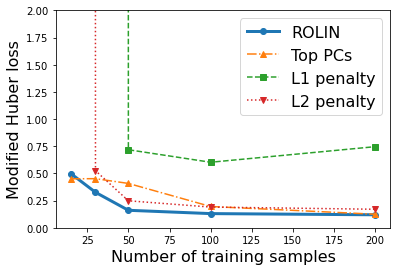}
      \caption{{\tt Htru} (p=8)}
      \label{fig:htru}
    \end{subfigure}
\end{figure}
\begin{figure}
\ContinuedFloat
    \begin{subfigure}{\textwidth}
      \centering
      \includegraphics[width=0.28\linewidth]{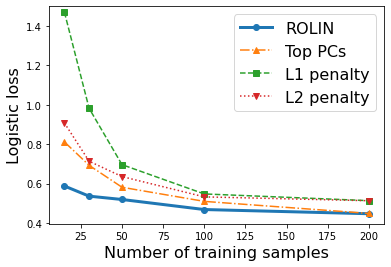}
      \includegraphics[width=0.28\linewidth]{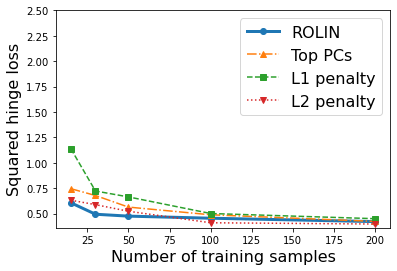}
      \includegraphics[width=0.28\linewidth]{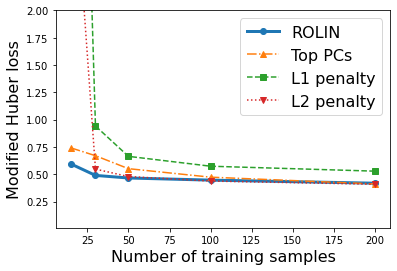}
      \caption{{\tt Phishing} (p=9)}
      \label{fig:phishing}
    \end{subfigure}
    \begin{subfigure}{\textwidth}
      \centering
      \includegraphics[width=0.28\linewidth]{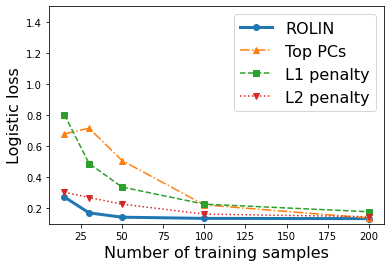}
      \includegraphics[width=0.28\linewidth]{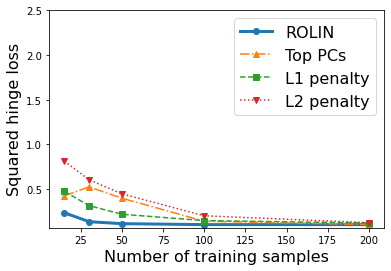}
      \includegraphics[width=0.28\linewidth]{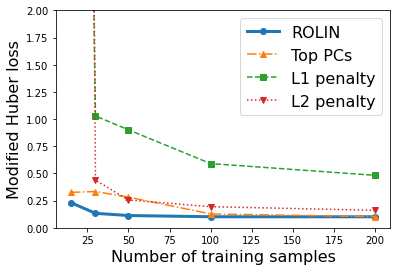}
      \caption{{\tt Breast cancer} (p=9)}
      \label{fig:breast_cancer}
    \end{subfigure}
    \begin{subfigure}{\textwidth}
      \centering
      \includegraphics[width=0.28\linewidth]{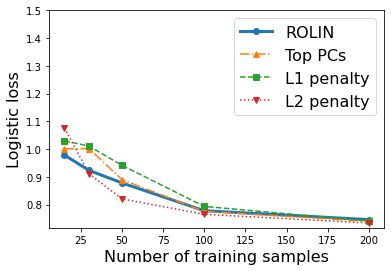}
      \includegraphics[width=0.28\linewidth]{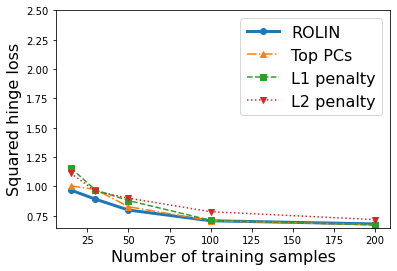}
      \includegraphics[width=0.28\linewidth]{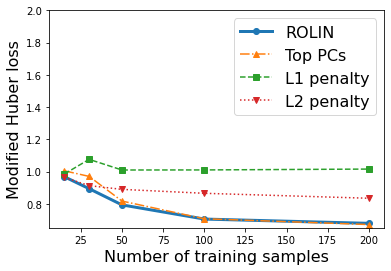}
      \caption{{\tt Magic} (p=9)}
      \label{fig:magic}
    \end{subfigure}
    \begin{subfigure}{\textwidth}
      \centering
      \includegraphics[width=0.28\linewidth]{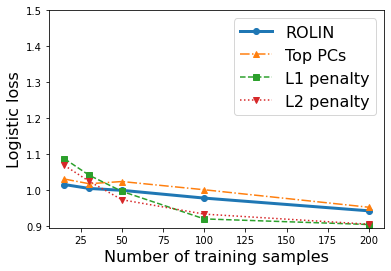}
      \includegraphics[width=0.28\linewidth]{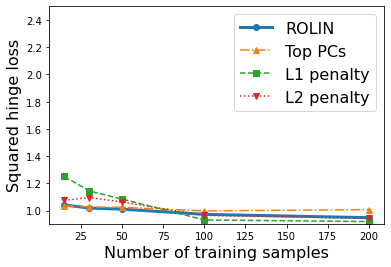}
      \includegraphics[width=0.28\linewidth]{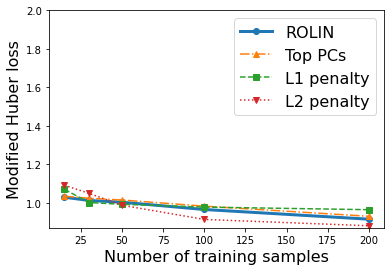}
      \caption{{\tt Avila} (p=10)}
      \label{fig:avila}
    \end{subfigure}
    \begin{subfigure}{\textwidth}
      \centering
      \includegraphics[width=0.28\linewidth]{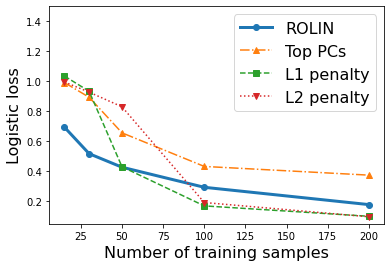}
      \includegraphics[width=0.28\linewidth]{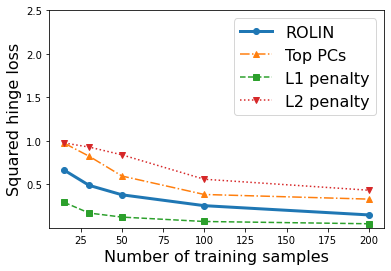}
      \includegraphics[width=0.28\linewidth]{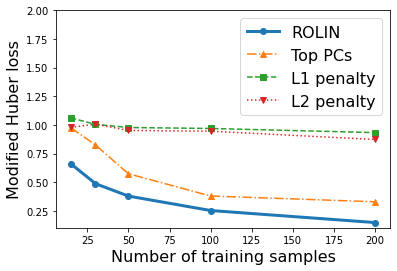}
      \caption{{\tt Electric} (p=12)}
      \label{fig:electric}
    \end{subfigure}
    \begin{subfigure}{\textwidth}
      \centering
      \includegraphics[width=0.28\linewidth]{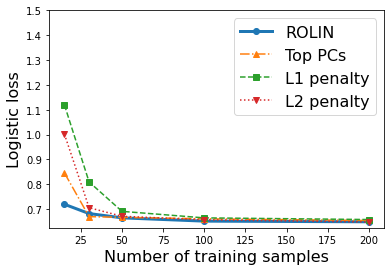}
      \includegraphics[width=0.28\linewidth]{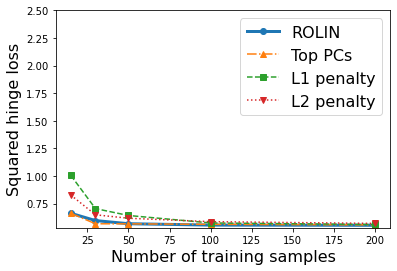}
      \includegraphics[width=0.28\linewidth]{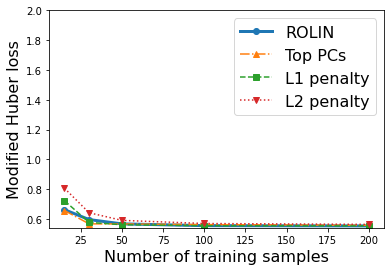}
      \caption{{\tt Music} (p=16)}
      \label{fig:music}
    \end{subfigure}
\end{figure}
\begin{figure}
\ContinuedFloat
    \begin{subfigure}{\textwidth}
      \centering
      \includegraphics[width=0.28\linewidth]{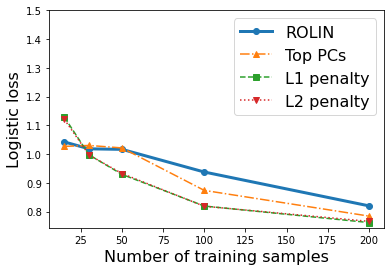}
      \includegraphics[width=0.28\linewidth]{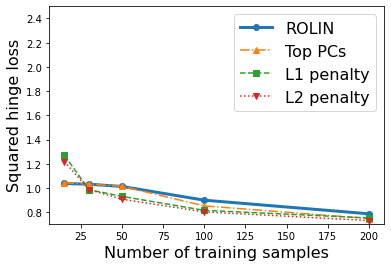}
      \includegraphics[width=0.28\linewidth]{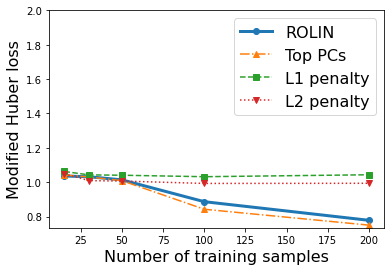}
      \caption{{\tt Diabetic} (p=19)}
      \label{fig:diabetic}
    \end{subfigure}
    \begin{subfigure}{\textwidth}
      \centering
      \includegraphics[width=0.28\linewidth]{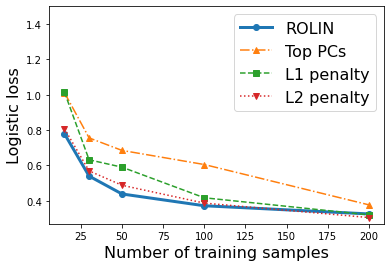}
      \includegraphics[width=0.28\linewidth]{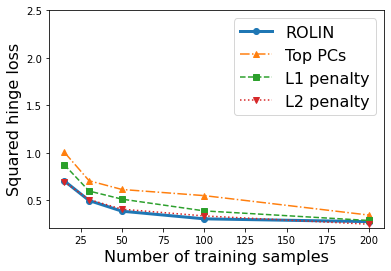}
      \includegraphics[width=0.28\linewidth]{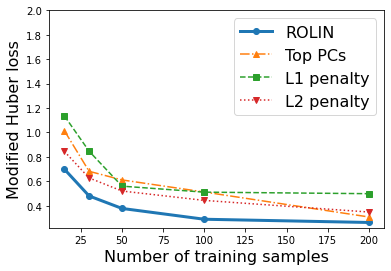}
      \caption{{\tt Frog} (p=22)}
      \label{fig:frog}
    \end{subfigure}
    \begin{subfigure}{\textwidth}
      \centering
      \includegraphics[width=0.28\linewidth]{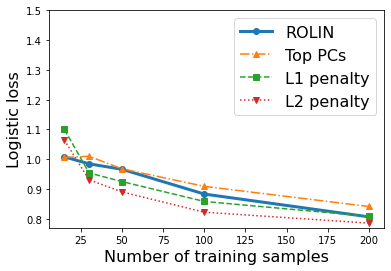}
      \includegraphics[width=0.28\linewidth]{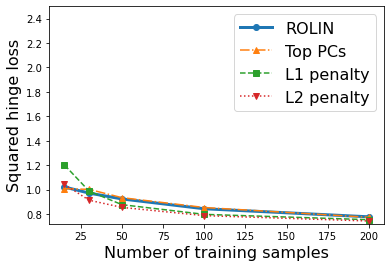}
      \includegraphics[width=0.28\linewidth]{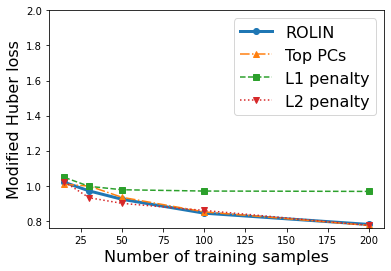}
      \caption{{\tt Sensor} (p=24)}
      \label{fig:sensor}
    \end{subfigure}
    \begin{subfigure}{\textwidth}
      \centering
      \includegraphics[width=0.28\linewidth]{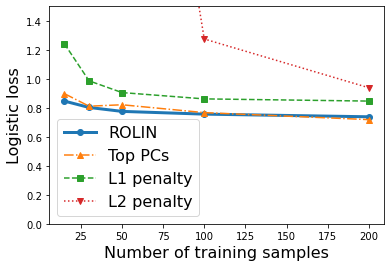}
      \includegraphics[width=0.28\linewidth]{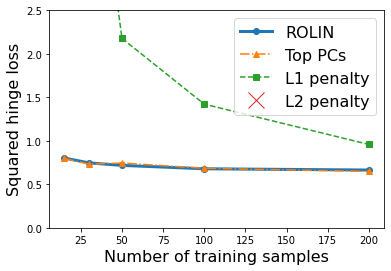}
      \includegraphics[width=0.28\linewidth]{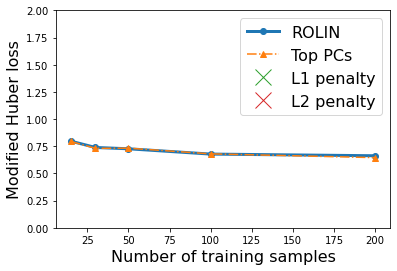}
      \caption{{\tt Credit} (p=26)}
      \label{fig:credit}
    \end{subfigure}
    \begin{subfigure}{\textwidth}
      \centering
      \includegraphics[width=0.28\linewidth]{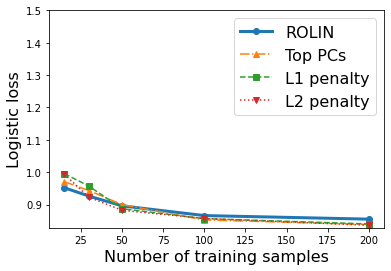}
      \includegraphics[width=0.28\linewidth]{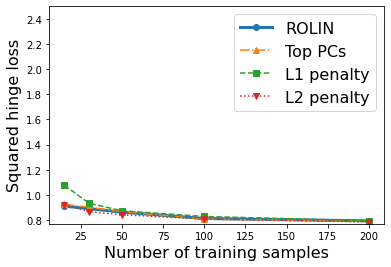}
      \includegraphics[width=0.28\linewidth]{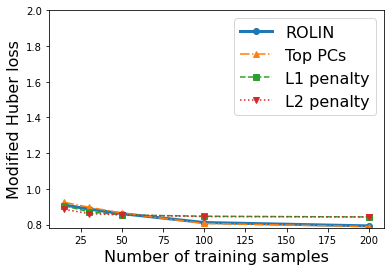}
      \caption{{\tt Gesture} (p=32)}
      \label{fig:gesture}
    \end{subfigure}
    \begin{subfigure}{\textwidth}
      \centering
      \includegraphics[width=0.28\linewidth]{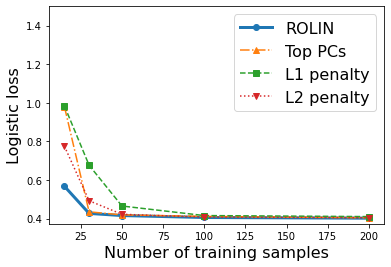}
      \includegraphics[width=0.28\linewidth]{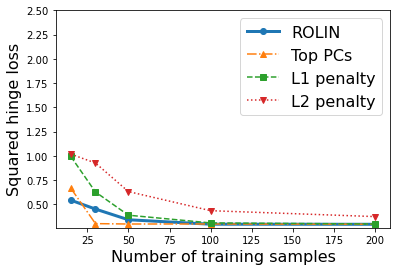}
      \includegraphics[width=0.28\linewidth]{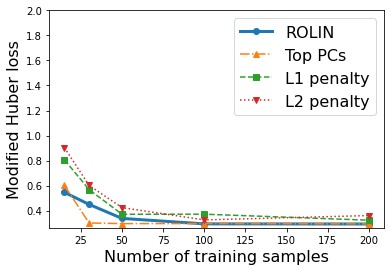}
      \caption{{\tt Theorem} (p=51)}
      \label{fig:theorem}
    \end{subfigure}
\end{figure}
\begin{figure}
\ContinuedFloat
    \begin{subfigure}{\textwidth}
      \centering
      \includegraphics[width=0.28\linewidth]{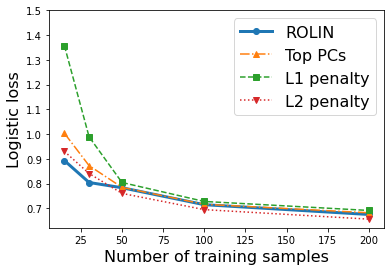}
      \includegraphics[width=0.28\linewidth]{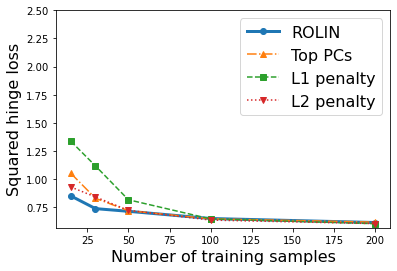}
      \includegraphics[width=0.28\linewidth]{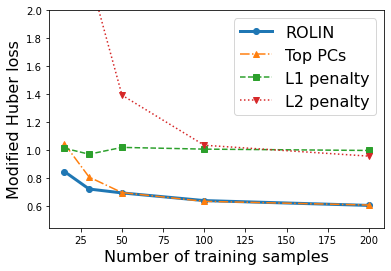}
      \caption{{\tt Objectivity} (p=59)}
      \label{fig:objectivity}
    \end{subfigure}
    \begin{subfigure}{\textwidth}
      \centering
      \includegraphics[width=0.28\linewidth]{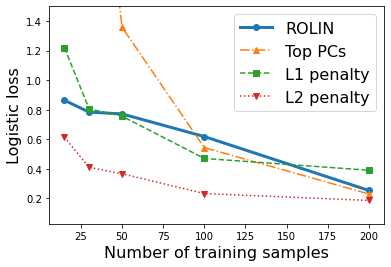}
      \includegraphics[width=0.28\linewidth]{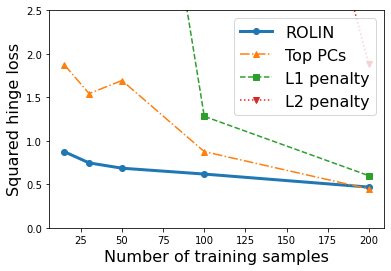}
      \includegraphics[width=0.28\linewidth]{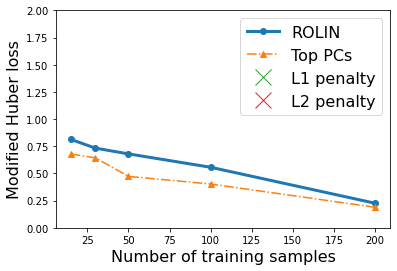}
      \caption{{\tt Buzz} (p=96)}
      \label{fig:buzz}
    \end{subfigure}
    \begin{subfigure}{\textwidth}
      \centering
      \includegraphics[width=0.28\linewidth]{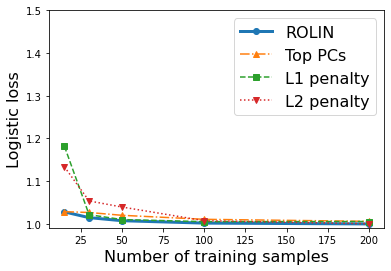}
      \includegraphics[width=0.28\linewidth]{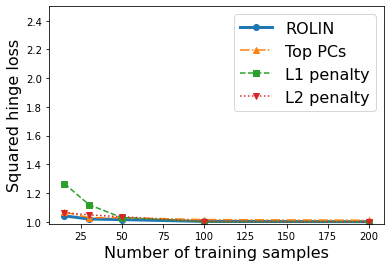}
      \includegraphics[width=0.28\linewidth]{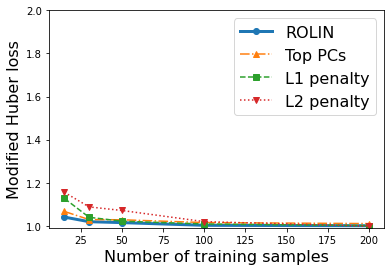}
      \caption{{\tt Defence of the Ancients game} (p=113)}
      \label{fig:dota}
    \end{subfigure}
    \begin{subfigure}{\textwidth}
      \centering
      \includegraphics[width=0.28\linewidth]{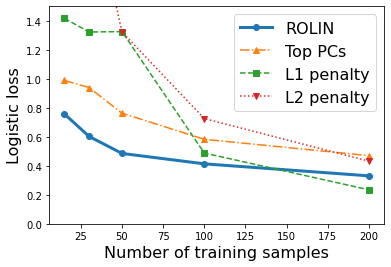}
      \includegraphics[width=0.28\linewidth]{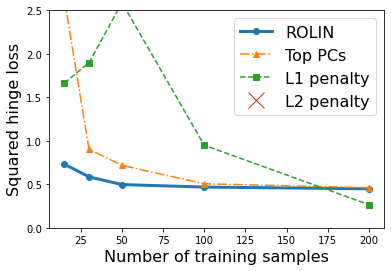}
      \includegraphics[width=0.28\linewidth]{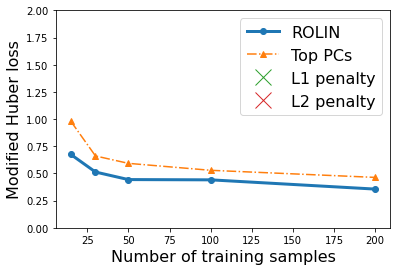}
      \caption{{\tt Gas sensor} (p=128)}
      \label{fig:gas_sensor}
    \end{subfigure}
    \begin{subfigure}{\textwidth}
      \centering
      \includegraphics[width=0.28\linewidth]{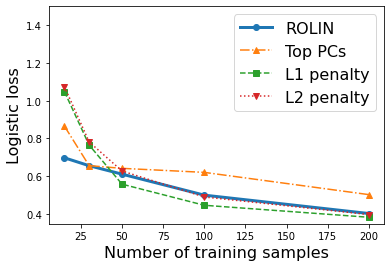}
      \includegraphics[width=0.28\linewidth]{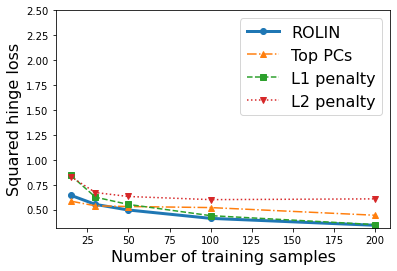}
      \includegraphics[width=0.28\linewidth]{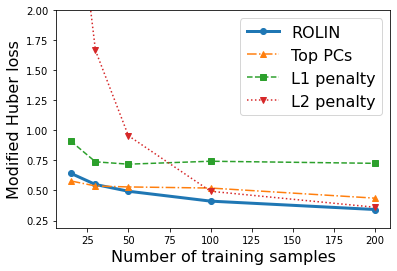}
      \caption{{\tt Musk} (p=166)}
      \label{fig:musk}
    \end{subfigure}
    \begin{subfigure}{\textwidth}
      \centering
      \includegraphics[width=0.28\linewidth]{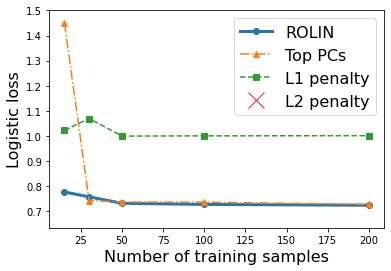}
      \includegraphics[width=0.28\linewidth]{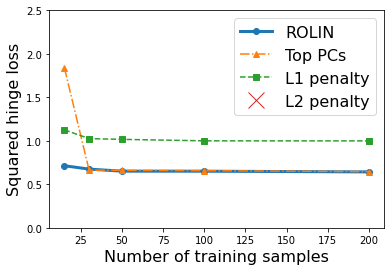}
      \includegraphics[width=0.28\linewidth]{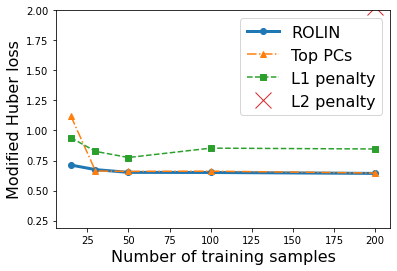}
      \caption{{\tt Seizure} (p=178)}
      \label{fig:seizure}
    \end{subfigure}
\end{figure}
\begin{figure}
\ContinuedFloat
    \begin{subfigure}{\textwidth}
      \centering
      \includegraphics[width=0.28\linewidth]{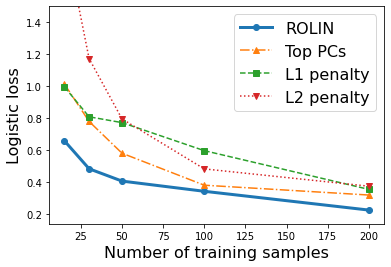}
      \includegraphics[width=0.28\linewidth]{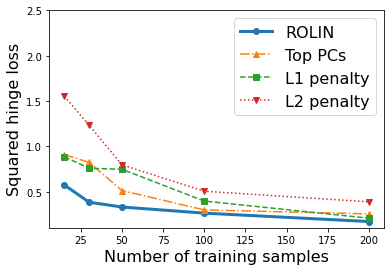}
      \includegraphics[width=0.28\linewidth]{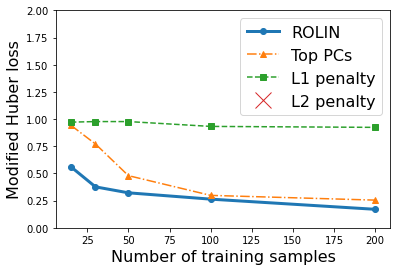}
      \caption{{\tt Facial} (p=300)}
      \label{fig:facial}
    \end{subfigure}
    \begin{subfigure}{\textwidth}
      \centering
      \includegraphics[width=0.28\linewidth]{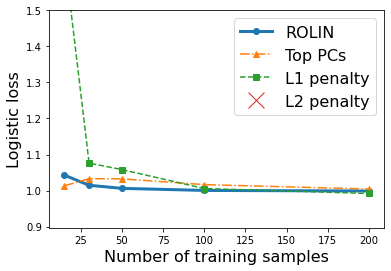}
      \includegraphics[width=0.28\linewidth]{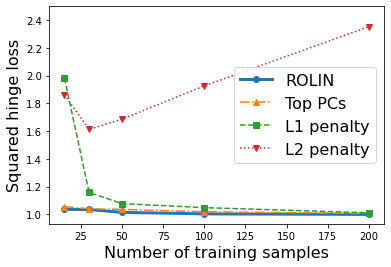}
      \includegraphics[width=0.28\linewidth]{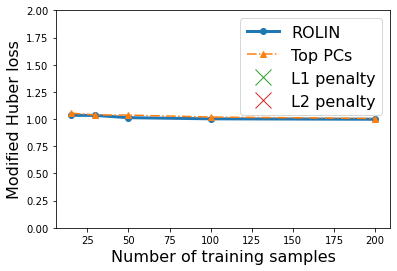}
      \caption{{\tt Fonts} (p=410)}
      \label{fig:fonts}
    \end{subfigure}
    \begin{subfigure}{\textwidth}
      \centering
      \includegraphics[width=0.28\linewidth]{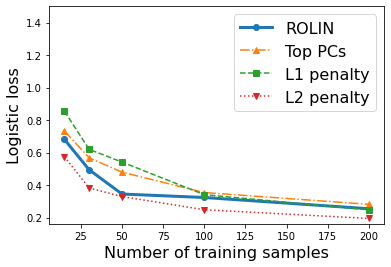}
      \includegraphics[width=0.28\linewidth]{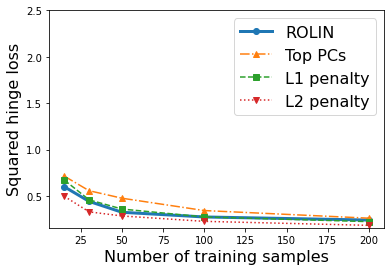}
      \includegraphics[width=0.28\linewidth]{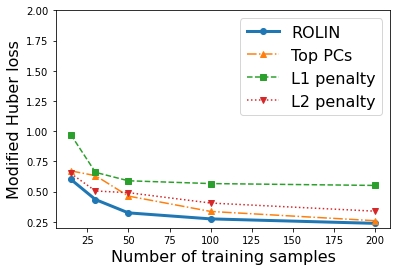}
      \caption{{\tt Activity} (p=561)}
      \label{fig:activity}
    \end{subfigure}
    \begin{subfigure}{\textwidth}
      \centering
      \includegraphics[width=0.28\linewidth]{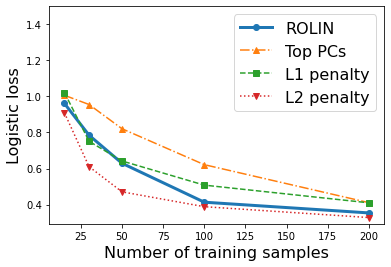}
      \includegraphics[width=0.28\linewidth]{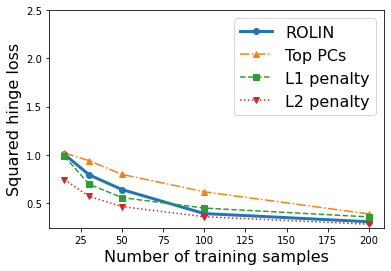}
      \includegraphics[width=0.28\linewidth]{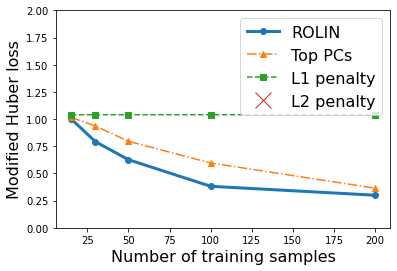}
      \caption{{\tt Gisette} (p=5000)}
      \label{fig:gisette}
    \end{subfigure}
    \begin{subfigure}{\textwidth}
      \centering
      \includegraphics[width=0.28\linewidth]{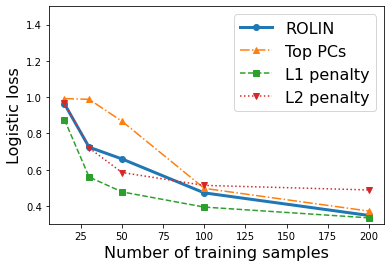}
      \includegraphics[width=0.28\linewidth]{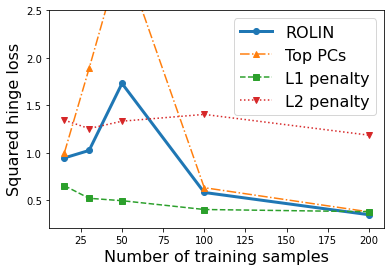}
      \includegraphics[width=0.28\linewidth]{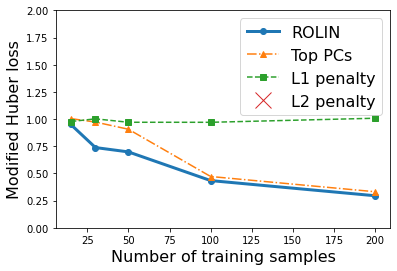}
      \caption{{\tt Hydraulic} (p=43680)}
      \label{fig:hydraulic}
    \end{subfigure}
\end{figure}

%% file: paper.bbl
\begin{thebibliography}{36}
\providecommand{\natexlab}[1]{#1}
\providecommand{\url}[1]{{#1}}
\providecommand{\urlprefix}{URL }
\expandafter\ifx\csname urlstyle\endcsname\relax
  \providecommand{\doi}[1]{DOI~\discretionary{}{}{}#1}\else
  \providecommand{\doi}{DOI~\discretionary{}{}{}\begingroup
  \urlstyle{rm}\Url}\fi
\providecommand{\eprint}[2][]{\url{#2}}

\bibitem[{Ben-Tal et~al.(2009)Ben-Tal, El~Ghaoui, and
  Nemirovski}]{ben2009robust}
Ben-Tal A, El~Ghaoui L, Nemirovski A (2009) Robust optimization

\bibitem[{Bhattacharyya(2004)}]{bhattacharyya_robust_2004}
Bhattacharyya C (2004) Robust classification of noisy data using second order
  cone programming approach. In: {Proceedings} of the International
  {Conference} on {Intelligent} {Sensing} and {Information} {Processing}, pp
  433--438

\bibitem[{Bi and Zhang(2004)}]{bi_support_2004}
Bi J, Zhang T (2004) Support {Vector} {Classification} with {Input} {Data}
  {Uncertainty}. Neural Information Processing Systems pp 161--168

\bibitem[{Blagus and Lusa(2013)}]{blagus_smote_2013}
Blagus R, Lusa L (2013) {SMOTE} for high-dimensional class-imbalanced data. BMC
  Bioinformatics 14(1):106

\bibitem[{Cover and Thomas(2006)}]{cover06elements}
Cover TM, Thomas JA (2006) Elements of Information Theory (Wiley Series in
  Telecommunications and Signal Processing). Wiley-Interscience, USA

\bibitem[{Davis and Kahan(1970)}]{davis1970rotation}
Davis C, Kahan WM (1970) The rotation of eigenvectors by a perturbation. {III}.
  SIAM Journal on Numerical Analysis 7(1):1--46

\bibitem[{De~Brabanter et~al.(2002)De~Brabanter, Pelckmans, Suykens, and
  Vandewalle}]{goos_robust_2002}
De~Brabanter J, Pelckmans K, Suykens JAK, Vandewalle J (2002) Robust
  {Cross}-{Validation} {Score} {Function} for {Non}-linear {Function}
  {Estimation}. In: Artificial {Neural} {Networks} — {ICANN}, vol 2415, pp
  713--719

\bibitem[{Delage and Ye(2010)}]{delage_distributionally_2010}
Delage E, Ye Y (2010) Distributionally {Robust} {Optimization} {Under} {Moment}
  {Uncertainty} with {Application} to {Data}-{Driven} {Problems}. Operations
  Research 58(3):595--612

\bibitem[{El~Ghaoui and Lebret(1997)}]{el1997robust}
El~Ghaoui L, Lebret H (1997) Robust solutions to least-squares problems with
  uncertain data. SIAM Journal on matrix analysis and applications
  18(4):1035--1064

\bibitem[{Erdoğan and Iyengar(2006)}]{erdogan_ambiguous_2006}
Erdoğan E, Iyengar G (2006) Ambiguous chance constrained problems and robust
  optimization. Mathematical Programming 107(1-2):37--61

\bibitem[{Feng et~al.(2014)Feng, Xu, Mannor, and Yan}]{feng_robust_2014}
Feng J, Xu H, Mannor S, Yan S (2014) Robust {Logistic} {Regression} and
  {Classification}. In: {Neural} {Information} {Processing} {Systems}, pp
  253--261

\bibitem[{{Gao Huang} et~al.(2012){Gao Huang}, {Shiji Song}, {Cheng Wu}, and
  {Keyou You}}]{gao_huang_robust_2012}
{Gao Huang}, {Shiji Song}, {Cheng Wu}, {Keyou You} (2012) Robust {Support}
  {Vector} {Regression} for {Uncertain} {Input} and {Output} {Data}. IEEE
  Transactions on Neural Networks and Learning Systems 23(11):1690--1700

\bibitem[{Globerson and Roweis(2006)}]{globerson06nightmare}
Globerson A, Roweis S (2006) Nightmare at test time: Robust learning by feature
  deletion. In: Proceedings of the 23rd International Conference on Machine
  Learning, ICML '06, pp 353--360

\bibitem[{Goh and Sim(2010)}]{goh_distributionally_2010}
Goh J, Sim M (2010) Distributionally {Robust} {Optimization} and {Its}
  {Tractable} {Approximations}. Operations Research 58(4-part-1):902--917

\bibitem[{Hastie et~al.(2009)Hastie, Tibshirani, and
  Friedman}]{hastie01statisticallearning}
Hastie T, Tibshirani R, Friedman J (2009) The Elements of Statistical Learning,
  2nd edn. Springer Series in Statistics, Springer New York Inc.

\bibitem[{Jiang and Guan(2016)}]{jiang_datadriven_2016}
Jiang R, Guan Y (2016) Data-driven chance constrained stochastic program.
  Mathematical Programming 158(1-2):291--327

\bibitem[{Jolliffe(1982)}]{jolliffe1982note}
Jolliffe IT (1982) A note on the use of principal components in regression.
  Applied Statistics pp 300--303

\bibitem[{Kordzakhia et~al.(2001)Kordzakhia, Mishra, and
  Reiersølmoen}]{kordzakhia_robust_2001}
Kordzakhia N, Mishra GD, Reiersølmoen L (2001) Robust estimation in the
  logistic regression model. Journal of Statistical Planning and Inference
  98(1):211--223

\bibitem[{Marcenko and Pastur(1967)}]{marcenko_distribution_1967}
Marcenko VA, Pastur LA (1967) Distribution of {Eigenvalues} for {Some} {Sets}
  of {Random} {Matrices}. Mathematics of the USSR-Sbornik 1(4):457--483

\bibitem[{Mohajerin~Esfahani and
  Kuhn(2018)}]{mohajerin_esfahani_datadriven_2018}
Mohajerin~Esfahani P, Kuhn D (2018) Data-driven distributionally robust
  optimization using the {Wasserstein} metric: performance guarantees and
  tractable reformulations. Mathematical Programming 171(1-2):115--166

\bibitem[{Mohri et~al.(2018)Mohri, Rostamizadeh, and
  Talwalkar}]{mohri12foundations}
Mohri M, Rostamizadeh A, Talwalkar A (2018) Foundations of Machine Learning,
  2nd edn

\bibitem[{Shafieezadeh-Abadeh et~al.(2015)Shafieezadeh-Abadeh, Esfahani, and
  Kuhn}]{shafi15distributionally}
Shafieezadeh-Abadeh S, Esfahani PM, Kuhn D (2015) Distributionally robust
  logistic regression. In: Neural Information Processing Systems, NIPS'15, pp
  1576--1584

\bibitem[{Shafieezadeh-Abadeh et~al.(2017)Shafieezadeh-Abadeh, Kuhn, and
  Esfahani}]{shafi17regularization}
Shafieezadeh-Abadeh S, Kuhn D, Esfahani PM (2017) Regularization via {Mass}
  {Transportation}. arXiv:171010016 [cs, math, stat] ArXiv: 1710.10016

\bibitem[{Shivaswamy et~al.(2006)Shivaswamy, Bhattacharyya, and
  Smola}]{shivaswamy_second_2006}
Shivaswamy PK, Bhattacharyya C, Smola AJ (2006) Second {Order} {Cone}
  {Programming} {Approaches} for {Handling} {Missing} and {Uncertain} {Data}.
  Journal of Machine Learning Research 7:1283--1314

\bibitem[{Sotiriou et~al.(2003)Sotiriou, Neo, McShane, Korn, Long, Jazaeri,
  Martiat, Fox, Harris, and Liu}]{sotiriou_breast_2003}
Sotiriou C, Neo SY, McShane LM, Korn EL, Long PM, Jazaeri A, Martiat P, Fox SB,
  Harris AL, Liu ET (2003) Breast cancer classification and prognosis based on
  gene expression profiles from a population-based study. Proceedings of the
  National Academy of Sciences 100(18):10393--10398

\bibitem[{Tibshirani and Manning(2014)}]{tibshirani_robust_2014}
Tibshirani J, Manning CD (2014) Robust {Logistic} {Regression} using {Shift}
  {Parameters}. In: Proceedings of the 52nd {Annual} {Meeting} of the
  {Association} for {Computational} {Linguistics} ({Volume} 2: {Short}
  {Papers}), pp 124--129

\bibitem[{Tibshirani(1996)}]{tibshirani_regression_1996}
Tibshirani R (1996) Regression {Shrinkage} and {Selection} via the {Lasso}.
  Journal of the Royal Statistical Society Series B (Methodological)
  58(1):267--288

\bibitem[{Trafalis and Gilbert(2006)}]{trafalis_robust_2006}
Trafalis TB, Gilbert RC (2006) Robust classification and regression using
  support vector machines. European Journal of Operational Research
  173(3):893--909

\bibitem[{Tzelepis et~al.(2018)Tzelepis, Mezaris, and
  Patras}]{tzelepis_linear_2018}
Tzelepis C, Mezaris V, Patras I (2018) Linear {Maximum} {Margin} {Classifier}
  for {Learning} from {Uncertain} {Data}. IEEE Transactions on Pattern Analysis
  and Machine Intelligence 40(12):2948--2962

\bibitem[{Wang and Pardalos(2014)}]{wang_survey_2014}
Wang X, Pardalos PM (2014) A {Survey} of {Support} {Vector} {Machines} with
  {Uncertainties}. Annals of Data Science 1(3-4):293--309

\bibitem[{Wiesemann et~al.(2014)Wiesemann, Kuhn, and
  Sim}]{wiesemann_distributionally_2014}
Wiesemann W, Kuhn D, Sim M (2014) Distributionally {Robust} {Convex}
  {Optimization}. Operations Research 62(6):1358--1376

\bibitem[{Wozabal(2012)}]{wozabal_framework_2012}
Wozabal D (2012) A framework for optimization under ambiguity. Annals of
  Operations Research 193(1):21--47

\bibitem[{Xu et~al.(2009{\natexlab{a}})Xu, Caramanis, and
  Mannor}]{xu2009robust}
Xu H, Caramanis C, Mannor S (2009{\natexlab{a}}) Robust regression and lasso.
  In: Neural Information Processing Systems, pp 1801--1808

\bibitem[{Xu et~al.(2009{\natexlab{b}})Xu, Caramanis, and
  Mannor}]{xu_robustness_2009}
Xu H, Caramanis C, Mannor S (2009{\natexlab{b}}) Robustness and
  {Regularization} of {Support} {Vector} {Machines}. Journal of Machine
  Learning Research 10:1485--1510

\bibitem[{Yu et~al.(2015)Yu, Wang, and Samworth}]{yu_useful_2015}
Yu Y, Wang T, Samworth RJ (2015) A useful variant of the {Davis}-{Kahan}
  theorem for statisticians. Biometrika 102(2):315--323

\bibitem[{Zhao et~al.(2019)Zhao, Chakrabarti, and Muthuraman}]{zhao19portfolio}
Zhao L, Chakrabarti D, Muthuraman K (2019) Portfolio construction by mitigating
  error amplification: The bounded-noise portfolio. Operations Research
  67(4):965--983

\end{thebibliography}
